\documentclass[10pt,journal,compsoc]{IEEEtran}

\ifCLASSOPTIONcompsoc
  \usepackage[nocompress]{cite}
\else
  \usepackage{cite}
\fi
\usepackage{url}
\usepackage{graphicx}
\usepackage{booktabs}
\usepackage{tabulary}

\usepackage{amsmath}
\usepackage{amssymb,amsthm,amsbsy}
\theoremstyle{plain}
\newtheorem{thm}{Theorem}%[section]

\ifCLASSOPTIONcompsoc
  \usepackage[caption=false,font=footnotesize,labelfont=sf,textfont=sf]{subfig}
\else
  \usepackage[caption=false,font=footnotesize]{subfig}
\fi

\begin{document}
\title{Training Deep Learning Based Denoisers without Ground Truth Data}

\author{Shakarim~Soltanayev,~%~\IEEEmembership{Student Member,~IEEE,}
        Se Young~Chun,~\IEEEmembership{Member,~IEEE} %membership info CHANGE
\IEEEcompsocitemizethanks
{\IEEEcompsocthanksitem S. Soltanayev and S.Y. Chun are with the Department
of Electrical Engineering, Ulsan National Institute of Science and Technology (UNIST), 50 UNIST-gil, Ulsan 44919, Republic of Korea. %\protect\\
E-mail: \{shakarim, sychun\}@unist.ac.kr}}%
%\thanks{Manuscript received - , 2019; revised -, 2019.}} %CHANGE

\markboth{Extended technical report of Soltanayev \& Chun, NeurIPS, 2018.}% CHANGE
{Soltanayev \MakeLowercase{\textit{et al.}}: Training and Refining Deep Learning Based Denoisers without Ground Truth Data}


\IEEEtitleabstractindextext{%
\begin{abstract}
Recently developed deep-learning-based denoisers often outperform state-of-the-art conventional denoisers such as the BM3D. 
They are typically trained to minimize the mean squared error (MSE) between the output image of a deep neural network (DNN) and a ground truth image.
Thus, it is important for deep-learning-based denoisers to use high quality noiseless ground truth data for high performance.
However, it is often challenging or even infeasible to obtain noiseless images
in some applications. Here,
we propose a method based on Stein's unbiased risk estimator (SURE) for training
DNN denoisers based only on the use of noisy images in the training data with Gaussian noise.
We demonstrate that our SURE-based method, without the use of ground truth data, is
able to train DNN denoisers to yield performances close to those networks trained with ground truth 
%, and to outperform the state-of-the-art denoiser BM3D 
for both grayscale and color images.
We also propose a SURE-based refining method with a noisy test image for further performance improvement. 
Our quick refining method outperformed conventional BM3D, deep image prior, and often the networks trained with ground truth.
%Our proposed achieved further improvements and outperformed conventional BM3D and recent deep image prior by large margins and often 
Potential extension of our SURE-based methods to Poisson noise model was also investigated.
%were achieved when %by including 
%noisy test images were used for refining of denoiser networks 
%using our proposed SURE-based method. 
%Code is available at https://github.com/Shakarim94/Net-SURE.
\end{abstract}

}

\maketitle

% \fi
\IEEEpeerreviewmaketitle

\IEEEraisesectionheading{\section{Introduction}\label{intro}}

\IEEEPARstart{D}{eep} neural network (DNN) has been successfully used in various high-level computer vision tasks, such as image classification~\cite{Krizhevsky:2012wl,He:2016ib,Szegedy:2017vxa}, object detection~\cite{Girshick:2014jx,Ren:2015ug,Redmon:2017gn}, semantic segmentation~\cite{chen14semantic,Long:2015cs,Chen:2018jn}, and image generation~\cite{Goodfellow:2014td,Kingma:2014tz,Zhu:2017hr}.
DNN has also been investigated for low-level computer vision tasks, such as  
image denoising~\cite{Vincent:2010vu,Burger:2012gm,Wang:2014cx,Zhang:2017eh,Lefkimmiatis:2017ka}, image inpainting~\cite{Xie:2012ws,Iizuka:2017bz,Yeh:2017ip}, image super resolution~\cite{dong2016image,Lim:2017it,Park:2018df}, and image restoration~\cite{Mao:2016ti,Gao:2017to,Zhang:2017vz}.
In particular, image denoising is a fundamental computer vision task that yields images with reduced noise, and improves the execution of other tasks, such as image classification~\cite{Vincent:2010vu} and image restoration~\cite{Zhang:2017vz}.

Deep learning based image denoisers
have yielded performances that are equivalent to or better than those of
conventional state-of-the-art denoising techniques such as BM3D~\cite{Dabov:2007fh}.
These deep denoisers typically train their networks by minimizing the mean-squared error (MSE)
between the output of a network and a ground truth (noiseless) image. 
Thus, it is crucial to have high quality noiseless images for high performance %deep learning denoisers.
denoising.
Thus far, DNN denoisers have been successful since 
high quality camera sensors and abundant light allow the acquisition of
%s to obtain massive amount of 
high quality, almost noiseless 2D images in daily environment tasks. 
Acquiring such high quality photographs is quite cheap %these days 
with the use of smart phones and digital cameras.

However, it is often challenging to apply currently developed DNN based image denoisers with minimum MSE to 
some application areas, such as hyperspectral remote sensing and medical imaging, where
the acquisition of noiseless ground truth data is expensive or sometimes even infeasible.
For example, hyperspectral imaging contains hundreds of spectral information per pixel, often leading to
%so that
increased noise in imaging sensors~\cite{Ye:2014du}.
Long acquisitions may improve image quality, but it is challenging to perform them with 
spaceborne or airborne imaging.
Similarly, in X-ray CT, %medical imaging, 
%ultra high resolution 3D MRI with sub-millimeter resolution often requires several hours of acquisition time for
%a single, high quality volume, but reducing acquisition time leads to increased noise.
%In X-ray CT, 
image noise can be substantially reduced by increasing the radiation dose.
Recent studies on deep learning based image denoisers
used CT images generated with normal doses as the ground truth so that 
denoising networks would be able to be trained to yield excellent performance~\cite{Chen:2017iy,Kang:2017jp}.
However, increased radiation dose 
leads to harmful effects in scanned subjects, while
excessively high doses may saturate the CT detectors.
%(e.g., in a similar manner to the acquisition of a photo of the sun without any filter).
Thus, acquiring ground truth data with newly developed scanners seems challenging without
compromising the subjects' safety.

Recently, deep learning based denoising methods have been investigated that do not require ground truth training images.
Noise2noise was developed for unsupervised training of DNNs for various applications including image denoising for Gaussian, Poisson, and Bernoulli noises~\cite{lehtinen2018noise2noise}.
%used in many applications including denoising
However, it requires %needs at least 
two noise realizations for each image for training. %, which is not always possible.
Deep image prior (DIP) does not require any training data and uses an input noisy image to perform blind denoising~\cite{ulyanov2017deep}. However,
it achieved lower performance than conventional BM3D and it required well-designed DNNs with hyperparameters and constraints to network architectures.
%only uses the noisy input image and  does not utilize any external training dataset.
%The method was successfully implemented for denoising of natural images, but achieved lower performance than BM3D.
%Moreover, many hyperparameters and constraints to network architectures are present.
We recently proposed a Stein's unbiased risk estimator (SURE) based unsupervised training method for DNN based Gaussian denoisers~\cite{NIPS2018_7587}.
Unlike noise2noise, it requires only one noise realizations for each image for training. %Moreover, unlike DIP, it can train DNN based denoisers and any DNNs can be used.

It is worth noting that SURE, an unbiased MSE estimator~\cite{Stein:1981vf}, has been investigated for optimizing the hyperparameters of conventional denoisers without ground truth data~\cite{donoho1995adapting,zhang1998adaptive}. The analytical form of SURE~\cite{Kocher:2009ix,Salmon:2010fa,Nguyen:2017iba}, a Monte-Carlo-based SURE (MC-SURE)~\cite{Ramani:2008ij}, and the approximation of a weak gradient of SURE~\cite{deledalle2014stein} have been investigated for the optimal or near-optimal denoisers with hyperparameters in low dimensional spaces.

In this paper, we propose SURE based unsupervised training and refining methods of DNN based denoisers by extending our previous work~\cite{NIPS2018_7587}
to 1) investigating unsupervised training of blind deep learning based denoisers with color images, 
2) developing a SURE based unsupervised refining method with an input test image for further performance improvement, 
and 3) investigating a training method of DNNs for Poisson noise instead of Gaussian %noise 
to demonstrate the feasibility of our proposed method for other noise type.
%We propose a SURE-based training method for deep neural network denoisers without ground truth data.
%a method to train  %As an initial step toward training deep learning based denoisers . % for medical imaging, we investigated In 
Section~\ref{background} reviews %, we review %key results elicited from 
SURE, MC-SURE, and supervised training of DNN denoisers. %Subsequently, i
Then, %in 
Section~\ref{approach} %, we 
describes proposed methods for unsupervised training and refining of blind DNN denoisers with color images %and 
for Gaussian noise and also investigates unsupervised training for Poisson noise.
% for unsupervised training of DNN denoisers without ground truth data for Gaussian noise, develop a training method to deal with color images and blind DNN denoisers, propose a unsupervised refining method with an input test image for performance boost, and investigate using MC-SURE and stochastic gradient descent for training deep learning based image denoisers. 
In Section~\ref{sim}, simulation results are presented to show state-of-the-art performance of our proposed methods over conventional methods as well as often over DNN based methods trained with ground truth.
%(a) conventional state-of-the-art denoiser (BM3D), (b) deep learning based denoiser trained with BM3D output as the ground truth, 
%(c) the same deep neural network denoiser with the proposed SURE training without the ground truth, and
%(d) the same denoiser network with ground truth data as a reference.
Lastly, Section~\ref{discuss} concludes this article %paper 
by discussing several potential issues for further studies.

%A preliminary version of this paper was published in~\cite{NIPS2018_7587}. Since then, our method has been used in compressive image recovery from undersampled measurements~\cite{zhussip2018simultaneous}.

\section{Background}
\label{background}

%In this section, major results of SURE~\cite{Stein:1981vf,Blu:2007fz} and
%Monte-Carlo SURE~\cite{Ramani:2008ij} 
%will be reviewed.
%stochastic approximation based optimization such as stochastic gradient descent
%(SGD)~\cite{Bottou:1998un} will be reviewed.

\subsection{Stein's unbiased risk estimator (SURE)}

A signal (or image) with Gaussian noise can be modeled as
\begin{equation}
	\boldsymbol{y} = \boldsymbol{x} + \boldsymbol{n}  
	\label{eq:signal}
\end{equation}
where $\boldsymbol{x} \in \mathbb{R}^K$ is an unknown signal in accordance with $\boldsymbol{x} \sim p(\boldsymbol{x})$,
$\boldsymbol{y} \in \mathbb{R}^K$ is a known measurement,
$\boldsymbol{n} \in \mathbb{R}^K$ is an \textit{i.i.d.} Gaussian noise 
such that $\boldsymbol{n} \sim \mathcal{N} (\boldsymbol{0}, \sigma^2 \boldsymbol{I})$, and where
$\boldsymbol{I}$ is an identity matrix. We denote $\boldsymbol{n} \sim \mathcal{N} (\boldsymbol{0}, \sigma^2 \boldsymbol{I})$
as $\boldsymbol{n} \sim \mathcal{N}_{0, \sigma^2}$.
An estimator of $\boldsymbol{x}$ from $\boldsymbol{y}$ (or denoiser) can be defined 
as a function $\boldsymbol{h}( \boldsymbol{y} )$ of $\boldsymbol{y}$ %such that
%\begin{equation}
%	\boldsymbol{h}( \boldsymbol{y} ) = \boldsymbol{y} + \boldsymbol{g} ( \boldsymbol{y} )
%	\label{eq:denoiser}
%\end{equation}
where $\boldsymbol{h}$ %, \boldsymbol{g}$ are functions 
is a function from $\mathbb{R}^K$ to $\mathbb{R}^K$. %Accordingly, 
The SURE for $\boldsymbol{h}( \boldsymbol{y} )$ can be derived as %follows,
\begin{equation}
\eta (\boldsymbol{h}( \boldsymbol{y} )) %&= 
%\sigma^2 + \frac{\| \boldsymbol{g} ( \boldsymbol{y} ) \|^2}{K} + \frac{2 \sigma^2}{K} \sum_{i=1}^K 
%\frac{\partial \boldsymbol{g}_i ( \boldsymbol{y} )}{\partial \boldsymbol{y}_i } \label{eq:sure} \\ &
=  \frac{\| \boldsymbol{y} - \boldsymbol{h} ( \boldsymbol{y} ) \|^2}{K}  - \sigma^2
+ \frac{2 \sigma^2}{K} \sum_{i=1}^K 
\frac{\partial \boldsymbol{h}_i ( \boldsymbol{y} )}{\partial \boldsymbol{y}_i }
%\nonumber
\label{eq:sure}
\end{equation}
where $\eta: \mathbb{R}^K \to \mathbb{R}$ and $\boldsymbol{y}_i$ is the $i$th element of $\boldsymbol{y}$.
For a fixed $\boldsymbol{x}$,
the following theorem holds:
\begin{thm}[\cite{Stein:1981vf,Blu:2007fz}]
	The random variable $\eta (\boldsymbol{h}( \boldsymbol{y} ))$ is an unbiased estimator of
	\[
	\mathrm{MSE}(\boldsymbol{h}( \boldsymbol{y} )) = \frac{1}{K} \| \boldsymbol{x} - \boldsymbol{h}( \boldsymbol{y} ) \|^2
	\]
	or
%	\begin{equation}
	\[
	\mathbb{E}_{\boldsymbol{n} \sim \mathcal{N}_{0, \sigma^2}} 
	\left\{ \frac{\| \boldsymbol{x} - \boldsymbol{h}( \boldsymbol{y} ) \|^2}{K}  \right\}
	= \mathbb{E}_{\boldsymbol{n} \sim \mathcal{N}_{0, \sigma^2}}  
	\left\{  \eta (\boldsymbol{h}( \boldsymbol{y} )) \right\}
	\]
%	\label{eq:expectsure}
%	\end{equation}
	\label{thm:sure}
\end{thm} 
\noindent where $\mathbb{E}_{\boldsymbol{n} \sim \mathcal{N}_{0, \sigma^2}}\{ \cdot \}$ is the expectation operator
in terms of the random vector $\boldsymbol{n}$. Note that in Theorem~\ref{thm:sure}, 
$\boldsymbol{x}$ is treated as a fixed, deterministic vector.

In practice, $\sigma^2$ can be estimated~\cite{Ramani:2008ij} and 
$\| \boldsymbol{y} - \boldsymbol{h} ( \boldsymbol{y} ) \|^2$
only requires the output of the estimator (or denoiser).
Unfortunately, it is challenging to calculate the last divergence term of (\ref{eq:sure}) analytically for general denoising methods $\boldsymbol{h}( \boldsymbol{y} )$.
%this term analytically for more general denoising methods. can be obtained 
%analytically in some special cases, such as in linear or NLM filters~\cite{Kocher:2009ix}.
%However, it is challenging to calculate this term analytically for more general denoising methods.

\subsection{Monte-Carlo Stein's unbiased risk estimator}

Ramani \textit{et al.} introduced a fast Monte-Carlo (MC) approximation
of the divergence term in (\ref{eq:sure}) for general denoisers~\cite{Ramani:2008ij}. For a fixed unknown
true image $\boldsymbol{x}$, the following theorem is valid:
\begin{thm}[{\cite{Ramani:2008ij}}] Let $\boldsymbol{\tilde{n}} \sim \mathcal{N}_{0, 1} \in \mathbb{R}^K$ 
	be independent of $\boldsymbol{n}$ and $\boldsymbol{y}$. Then,
	\begin{equation}
	\sum_{i=1}^K \frac{\partial \boldsymbol{h}_i ( \boldsymbol{y} )}{\partial \boldsymbol{y}_i } = 
	\lim_{\epsilon \to 0} \mathbb{E}_{\boldsymbol{\tilde{n}}} \left\{ \boldsymbol{\tilde{n}}^\mathrm{t}  
	\left(  \frac{\boldsymbol{h} ( \boldsymbol{y} + \epsilon \boldsymbol{\tilde{n}} ) - \boldsymbol{h} ( \boldsymbol{y} )}{\epsilon} \right)\right\},
	\end{equation}
	provided that $\boldsymbol{h} ( \boldsymbol{y} )$ admits a well-defined second-order Taylor expansion. 
	If not, this is still valid in the weak sense provided that $\boldsymbol{h} ( \boldsymbol{y} )$ is tempered.
	\label{thm:mcsure}
\end{thm}
\noindent Based on the Theorem~\ref{thm:mcsure}, the divergence term in (\ref{eq:sure}) can be approximated by 
one realization of $\boldsymbol{\tilde{n}} \sim \mathcal{N}_{0, 1}$ and a fixed small positive value $\epsilon$:
\begin{equation}
\frac{1}{K} \sum_{i=1}^K \frac{\partial \boldsymbol{h}_i ( \boldsymbol{y} )}{\partial \boldsymbol{y}_i } \approx
\frac{1}{\epsilon K} \boldsymbol{\tilde{n}}^\mathrm{t}  
\left(  \boldsymbol{h} ( \boldsymbol{y} + \epsilon \boldsymbol{\tilde{n}} ) - \boldsymbol{h} ( \boldsymbol{y} ) \right),
\label{eq:approx}
\end{equation}
where $\mathrm{t}$ is the transpose operator.
This expression has been shown to yield accurate unbiased estimates of MSE for many conventional denoising
methods $\boldsymbol{h} ( \boldsymbol{y} )$~\cite{Ramani:2008ij}.

\subsection{Supervised training of DNN based denoisers}

A typical risk for image denoisers with the signal generation model (\ref{eq:signal}) is
\begin{equation}
\mathbb{E}_{\boldsymbol{x} \sim p(\boldsymbol{x}), \boldsymbol{n} \sim \mathcal{N}_{0, \sigma^2} }
\| \boldsymbol{x} - \boldsymbol{h} ( \boldsymbol{y};  {\boldsymbol \theta} ) \|^2,
\label{eq:riskdenoiser}
\end{equation}
where $\boldsymbol{h} ( \boldsymbol{y};  {\boldsymbol \theta} )$ is a deep learning based denoiser parametrized 
with a large-scale vector ${\boldsymbol \theta} \in \mathbb{R}^P$.
It is usually infeasible to calculate (\ref{eq:riskdenoiser}) exactly due to expectation operator.
Thus, the empirical risk for (\ref{eq:riskdenoiser}) is used as a cost function: % as follows:
\begin{equation}
\frac{1}{N} \sum_{j=1}^N
\| \boldsymbol{h} ( \boldsymbol{y}^{(j)};  {\boldsymbol \theta} )  - \boldsymbol{x}^{(j)} \|^2,
\label{eq:riskdenoiser_emp}
\end{equation}
where $\{ (\boldsymbol{x}^{(1)}, \boldsymbol{y}^{(1)}), \cdots, (\boldsymbol{x}^{(N)}, \boldsymbol{y}^{(N)}) \}$ are the $N$ pairs of a training dataset, sampled from the joint distribution of 
$\boldsymbol{x}^{(j)} \sim p(\boldsymbol{x})$ and $\boldsymbol{n}^{(j)} \sim \mathcal{N}_{0, \sigma^2}$. 
Note that (\ref{eq:riskdenoiser_emp}) is an unbiased estimator of (\ref{eq:riskdenoiser}).

To train the deep learning network $\boldsymbol{h} ( \boldsymbol{y};  {\boldsymbol \theta} )$ with respect to ${\boldsymbol \theta}$,
a gradient-based optimization algorithm is used such as the stochastic gradient descent (SGD)~\cite{Bottou:1998un},
momentum, Nesterov momentum~\cite{nesterov1983method}, or 
the Adam optimization algorithm~\cite{Kingma:2015us}. For any gradient-based optimization method, it is essential to calculate
the gradient of (\ref{eq:riskdenoiser}) with respect to ${\boldsymbol \theta}$: % as follows:
\begin{equation}
\mathbb{E}_{\boldsymbol{x} \sim p(\boldsymbol{x}), \boldsymbol{n} \sim \mathcal{N}_{0, \sigma^2} }
2 \nabla_{\boldsymbol \theta}  \boldsymbol{h} ( \boldsymbol{y};  {\boldsymbol \theta} )^\mathrm{t} \left( \boldsymbol{h} ( \boldsymbol{y};  {\boldsymbol \theta} )  - \boldsymbol{x} \right).
\label{eq:graddenoiser}
\end{equation}
Therefore, it is sufficient to calculate the gradient of the empirical risk (\ref{eq:riskdenoiser_emp}) to approximate
(\ref{eq:graddenoiser}). % for any gradient-based optimization.

In practice, calculating the gradient of (\ref{eq:riskdenoiser_emp}) for large $N$ is inefficient since 
a small amount of well-shuffled training data can often well-approximate the gradient of (\ref{eq:riskdenoiser_emp}).
Thus, a mini-batch is typically used for efficient DNN training by calculating the mini-batch empirical risk as follows:
\begin{equation}
\frac{1}{M} \sum_{j=1}^M
\| \boldsymbol{h} ( \boldsymbol{y}^{(j)};  {\boldsymbol \theta} )  - \boldsymbol{x}^{(j)} \|^2,
\label{eq:riskdenoiser_emp_batch}
\end{equation}
%where $M$ is the number of one mini-match.
where $M$ is the size of one mini-batch.
Equation (\ref{eq:riskdenoiser_emp_batch}) is still an unbiased estimator of (\ref{eq:riskdenoiser})
provided that the training data is randomly permuted every epoch. % and the same data is used
%no more than once per epoch.
In deep learning based image processing such as image denoising or single image super resolution, it is often more efficient to use image patches instead of
whole images for training. For example, $\boldsymbol{x}^{(j)}$ and $\boldsymbol{y}^{(j)}$ can be image patches from a ground truth image and a noisy image, respectively.

\section{Methods}
\label{approach}

We will develop our proposed MC-SURE-based method for training and fine-tuning
deep learning based (blind) denoisers without ground truth images by assuming a Gaussian
noise model in (\ref{eq:signal}). We will also investigate how to extend it to other noise model such as a Poisson model.

\subsection{Unsupervised training of DNN based denoisers}

To incorporate MC-SURE into a stochastic gradient-based optimization algorithm for training, such as the SGD or the Adam optimization algorithms, % for training, 
%the stochastic approximation method for training, 
we modify the risk (\ref{eq:riskdenoiser}) in accordance with %as
\begin{equation}
\mathbb{E}_{\boldsymbol{x} \sim p(\boldsymbol{x}) } \left[
\mathbb{E}_{\boldsymbol{n} \sim \mathcal{N}_{0, \sigma^2} } \left(
\| \boldsymbol{x} - \boldsymbol{h} ( \boldsymbol{y};  {\boldsymbol \theta} ) \|^2 |  \boldsymbol{x} \right)  \right],
\label{eq:riskdenoiser_mod}
\end{equation}
where (\ref{eq:riskdenoiser_mod}) is equivalent to (\ref{eq:riskdenoiser}) owing to conditioning.

From Theorem~\ref{thm:sure}, an unbiased estimator for 
$\mathbb{E}_{\boldsymbol{n} \sim \mathcal{N}_{0, \sigma^2} } \left(
\| \boldsymbol{x} - \boldsymbol{h} ( \boldsymbol{y};  {\boldsymbol \theta} ) \|^2 |  \boldsymbol{x} \right)$ can be derived as
\[
K \eta (\boldsymbol{h}( \boldsymbol{y}; {\boldsymbol \theta} ))
\]
such that for a fixed $\boldsymbol{x}$,
\begin{align}
&\mathbb{E}_{\boldsymbol{n} \sim \mathcal{N}_{0, \sigma^2} } \left(
\| \boldsymbol{x} - \boldsymbol{h} ( \boldsymbol{y};  {\boldsymbol \theta} ) \|^2 |  \boldsymbol{x} \right) \\
&= \mathbb{E}_{\boldsymbol{n} \sim \mathcal{N}_{0, \sigma^2} } 
\| \boldsymbol{x} - \boldsymbol{h} ( \boldsymbol{y};  {\boldsymbol \theta} ) \|^2
= K  \mathbb{E}_{\boldsymbol{n} \sim \mathcal{N}_{0, \sigma^2} } 
\eta (\boldsymbol{h}( \boldsymbol{y}; {\boldsymbol \theta} )). \nonumber
\end{align}
Therefore, a new risk for image denoisers will be
\begin{equation}
\mathbb{E}_{\boldsymbol{x}, \boldsymbol{n} \sim p(\boldsymbol{x}), \mathcal{N}_{0, \sigma^2} }
\| \boldsymbol{x} - \boldsymbol{h} ( \boldsymbol{y};  {\boldsymbol \theta} ) \|^2
= \mathbb{E}_{\boldsymbol{y} \sim q(\boldsymbol{y}) } K \eta (\boldsymbol{h}( \boldsymbol{y}; {\boldsymbol \theta} ))
\label{eq:riskdenoiser2}
\end{equation}
where $q(\boldsymbol{y}) = p(\boldsymbol{x}) \mathcal{N}_{0, \sigma^2}$ due to (\ref{eq:signal}).
Note that 
%where there is 
no noiseless ground truth data $\boldsymbol{x}$ is required for the right-side risk function of (\ref{eq:riskdenoiser2}) and there is no approximation
in (\ref{eq:riskdenoiser2}).

The strong law of large numbers (SLLN) allows (\ref{eq:riskdenoiser2}) to be well-approximated as the following empirical risk that is
%Thus, using the empirical risk expression in (\ref{eq:riskdenoiser_emp_batch}), 
an unbiased estimator for (\ref{eq:riskdenoiser}):
\begin{align}
&\frac{1}{M} \sum_{j=1}^M \bigg\{
\| \boldsymbol{y}^{(j)} - \boldsymbol{h} ( \boldsymbol{y}^{(j)}; {\boldsymbol \theta} ) \|^2 - K \sigma^2 \nonumber \\
&+ 2 \sigma^2 \sum_{i=1}^K 
\frac{\partial \boldsymbol{h}_i ( \boldsymbol{y}^{(j)}; {\boldsymbol \theta} )}{\partial \boldsymbol{y}_i } \bigg\}.
\label{eq:riskdenoiser_emp_batch_prop}
\end{align}
Finally, the last divergence term in (\ref{eq:riskdenoiser_emp_batch_prop}) can be approximated
using MC-SURE so that the final %unbiased risk 
estimator for (\ref{eq:riskdenoiser}) will be
\begin{align}
&\frac{1}{M} \sum_{j=1}^M \bigg\{
\| \boldsymbol{y}^{(j)} - \boldsymbol{h} ( \boldsymbol{y}^{(j)}; {\boldsymbol \theta} ) \|^2 - K \sigma^2  \label{eq:riskdenoiser_emp_batch_prop_mcsure} \\ 
&+ \frac{2 \sigma^2}{\epsilon} (\boldsymbol{\tilde{n}}^{(j)})^\mathrm{t}  
\left(  \boldsymbol{h} ( \boldsymbol{y}^{(j)} + \epsilon \boldsymbol{\tilde{n}}^{(j)} ; {\boldsymbol \theta}) - \boldsymbol{h} ( \boldsymbol{y}^{(j)} ; {\boldsymbol \theta}) \right) \bigg\},
\nonumber
\end{align}
where $\epsilon$ is a small fixed positive number and $\boldsymbol{\tilde{n}}^{(j)}$ is a single realization from the 
standard normal distribution for each training data $j$.
In order to make sure that the estimator (\ref{eq:riskdenoiser_emp_batch_prop_mcsure}) is unbiased,
the order of $\boldsymbol{y}^{(j)}$ should be randomly permuted and the new set of $\boldsymbol{\tilde{n}}^{(j)}$
should be generated at every epoch. 

We propose to use MC based divergence approximation instead of direct calculation for it mainly due to computation 
complexity. Note the the last term of (\ref{eq:riskdenoiser_emp_batch_prop}) contains $MK$ derivatives of the denoiser with respect to image intensity on each pixel.
Moreover, during the training, the derivative with respect to ${\boldsymbol \theta}$ must be calculated in each 
iteration %epoch 
so that the total of $KMP$ derivatives must be evaluated for one mini-batch
iteration %epoch 
(e.g., $K = 50^2 = 2500$ in one of our simulations). However, our MC based approximation allows to reduce computation up to $2MP$.

The deep learning based image denoiser with the cost function of 
(\ref{eq:riskdenoiser_emp_batch_prop_mcsure}) can be implemented using a deep learning development framework,
such as TensorFlow~\cite{Abadi:2016:TSL:3026877.3026899}, by properly defining the cost function.
Thus, the gradient of (\ref{eq:riskdenoiser_emp_batch_prop_mcsure}) can be automatically calculated
when the training is performed.

One of the potential advantages of our SURE-based training method is that we can use all the available data without
noiseless ground truth images. In other words, we can train deep neural networks
with the use of both training and testing data.
%not only with training data, but also with test data together. 
This advantage may further improve the performance of deep learning based denoisers.
We will investigate more about this shortly.

Lastly, almost any DNN denoiser 
can utilize our MC-SURE-based training by modifying
the cost function from (\ref{eq:riskdenoiser_emp_batch}) to (\ref{eq:riskdenoiser_emp_batch_prop_mcsure})
as far as it satisfies the condition in Theorem~\ref{thm:mcsure}. 
Many deep learning based denoisers with differentiable activation functions ({e.g.}, sigmoid)
can comply with this condition.
Some denoisers with piecewise differentiable activation functions ({e.g.}, ReLU)
can still utilize Theorem~\ref{thm:mcsure} in the weak sense since
\[
\| \boldsymbol{h} ( \boldsymbol{y}; {\boldsymbol \theta} ) \| \le C_0 (1 + \| \boldsymbol{y} \|^{n_0} ),
\]
for some $n_0 > 1$ and $C_0 > 0$. Therefore, we expect that our proposed method 
should work for most deep learning image 
denoisers~\cite{Vincent:2010vu,Burger:2012gm,Wang:2014cx,Zhang:2017eh,Lefkimmiatis:2017ka}.
In our simulations, we demonstrate that our methods work for~\cite{Vincent:2010vu,Zhang:2017eh}.

\subsection{Unsupervised training of blind image denoisers}

DNN based denoisers are often trained as blind image denoisers for a certain range of noise levels instead of for one noise level.
It seems straightforward to extend (\ref{eq:riskdenoiser_emp_batch_prop_mcsure}) to the case of blind denoising. Assuming that an image (or image patch)
contains noise with the level of $\sigma^{(j)}$, we derived the following empirical risk:
\begin{align}
&\frac{1}{M} \sum_{j=1}^M \bigg\{
\| \boldsymbol{y}^{(j)} - \boldsymbol{h} ( \boldsymbol{y}^{(j)}; {\boldsymbol \theta} ) \|^2 - K \left( \sigma^{(j)} \right)^2 \label{eq:riskdenoiser_emp_batch_prop_mcsure_multi} \\ 
&+ \frac{2 \left( \sigma^{(j)} \right)^2 }{\epsilon^{(j)}} (\boldsymbol{\tilde{n}}^{(j)})^\mathrm{t}  
\left(  \boldsymbol{h} ( \boldsymbol{y}^{(j)} + \epsilon \boldsymbol{\tilde{n}}^{(j)} ; {\boldsymbol \theta}) - \boldsymbol{h} ( \boldsymbol{y}^{(j)} ; {\boldsymbol \theta}) \right) \bigg\}
\nonumber
\end{align}
where $\epsilon^{(j)}$ indicates that it can vary according to the noise level $\sigma^{(j)}$.
Note that the SLLN for (\ref{eq:riskdenoiser_emp_batch_prop_mcsure_multi}) holds if 
\[
\sum_{j=1}^\infty {\left( \sigma^{(j)} \right)^2} / {j^2}
\]
converges. Thus, any blind denoiser that has a finite range of noise levels should meet this sufficient condition.

\subsection{Unsupervised refining of DNN based denoisers}

One of the advantages for our proposed method is its ability to refining a trained DNN based denoisers using noisy test image(s).
Assume that there is a trained DNN based denoiser $\boldsymbol{h} ( \cdot ; \boldsymbol{\hat{\theta}} )$ where $\boldsymbol{\hat{\theta}}$ is a trained DNN parameter vector.
For a test image to denoise $\boldsymbol{\tilde{y}}$, we propose to refine $\boldsymbol{h} ( \cdot ; \boldsymbol{\hat{\theta}} )$ by the following minimization:
\begin{align}
&\min_{\boldsymbol{\theta}, \boldsymbol{B}\boldsymbol{\theta} = \boldsymbol{B} \boldsymbol{\hat{\theta}}}  \bigg\{
\| \boldsymbol{\tilde{y}}  - \boldsymbol{h} ( \boldsymbol{\tilde{y}} ; {\boldsymbol \theta} ) \|^2 - K \tilde{\sigma}^2 \label{eq:riskdenoiser_refine} \\ 
&+ \frac{2 \tilde{\sigma}^2 }{\tilde{\epsilon}} \boldsymbol{\tilde{n}}^\mathrm{t}  
\left(  \boldsymbol{h} ( \boldsymbol{\tilde{y}} + \tilde{\epsilon} \boldsymbol{\tilde{n}} ; {\boldsymbol \theta}) - \boldsymbol{h} ( \boldsymbol{\tilde{y}} ; {\boldsymbol \theta}) \right) \bigg\}
\nonumber
\end{align}
where $\boldsymbol{B}$ is the mask to select parameters in $\boldsymbol{h} ( \cdot ; \boldsymbol{\theta} )$ that
one would like to fix (e.g., batch normalization layers).
%where $\boldsymbol{B}$ is the mask to select parameters in all batch normalization layers of $\boldsymbol{h} ( \cdot ; \boldsymbol{\theta} )$.

\subsection{Unsupervised training for Poisson noise}

In this section we extend our proposed MC-SURE based training method to Poisson noise case. A more general
mixed Poisson-Gaussian noise model is given as follows:

\begin{equation*}
\boldsymbol{y} = \zeta \boldsymbol{z} + \boldsymbol{n} \;\;
\text{with} \;\;
\begin{cases}
\boldsymbol{ z} \sim  \mathcal{P} \left(  {\boldsymbol{x}} / {\zeta} \right) \\
\boldsymbol{n} \sim \mathcal{N} (\boldsymbol{0}, \sigma^2 \boldsymbol{I})
\end{cases}
\label{eq:mpg_signal}
\end{equation*}
where $\boldsymbol{z} \in \mathbb{R}^K$ is a random variable that follows a Poisson distribution and $\zeta \geq 0 $
is the gain of the acquisition process. 

Luisier \textit{et al.}~\cite{luisier2011image} proposed an unbiased estimator of the MSE called PURE 
for the case of $\zeta = 1$. In~\cite{le2014unbiased}, a more general PG-URE estimator and its Monte-Carlo 
approximation were presented. 

By %We can 
utilizing them for the case of pure Poisson noise (i.e., in the absence of additive Gaussian noise), we derived an %to get the 
unbiased risk estimator for (\ref{eq:riskdenoiser}) for Poisson denoising: % as follows:
\begin{align}
&\frac{1}{M} \sum_{j=1}^M \bigg\{
\| \boldsymbol{y}^{(j)} - \boldsymbol{h} ( \boldsymbol{y}^{(j)}; {\boldsymbol \theta} ) \|^2 - \zeta\sum_{i=1}^K \boldsymbol{y}^{(j)}_i 
\nonumber \\
&+ \frac{2 \zeta}{\dot{\epsilon}}
(\boldsymbol{\dot{n}^{(j)}} \odot \boldsymbol{y^{(j)}})^t 
\left( \boldsymbol{h} ( \boldsymbol{y^{(j)}} + \dot{\epsilon} \boldsymbol{\dot{n}^{(j)}} ) - \boldsymbol{h} ( \boldsymbol{y^{(j)}} ) \right),
\label{eq:riskdenoiser_emp_batch_prop_mcpure}
\end{align}
where $\boldsymbol{\dot{n}} \in \mathbb{R}^K$ is a random variable following a binary distribution taking values
$-1$ and $-1$ with probability $0.5$ each, $\dot{\epsilon}$ is a small positive number similar to $\epsilon$ from 
(\ref{eq:approx}), and $\odot$ is a componentwise multiplication.

\section{Simulation results}
\label{sim}

In this section, denoising simulation results are presented with the MNIST dataset
using a simple stacked denoising autoencoder (SDA)~\cite{Vincent:2010vu},
%,  CIFAR-10 dataset using 
%a convolution-deconvolution deep network (U-NET)~\cite{ronneberger2015u},
and a large-scale natural image dataset using a 
deep convolutional neural network (CNN) image denoiser (DnCNN)~\cite{Zhang:2017eh}.
%are presented. 
%For both cases, conventional state-of-the-art image denoiser, BM3D~\cite{Dabov:2007fh},
%(does not require any training)
%was also tested.

All of the networks presented in this section 
(denoted by NET, which can be either %one among 
SDA or
% , U-NET,and 
DnCNN)
were trained using one of the following two optimization objectives: 
(MSE) the minimum MSE between a denoised image and its ground truth image in (\ref{eq:riskdenoiser_emp_batch}) and  
(SURE) our proposed minimum MC-SURE without ground truth in (\ref{eq:riskdenoiser_emp_batch_prop_mcsure}). 
%Note that ground truth images were not used in the latter case. % we did not use 
NET-MSE methods generated noisy training images at every epochs in accordance with~\cite{Zhang:2017eh}, while
our proposed NET-SURE methods used only noisy images %only that were 
obtained before training.
%Optimizing MSE training images were corrupted by different realizations of Gaussian noise for each epochWhen 
%while for SURE optimization training images were corrupted only once at the start of the training. 
%We also propose the SURE-T method which utilized noisy test images %along 
We also propose the SURE-FT method which utilized noisy test images for fine-tuning (refining) pretrained networks.
%This method is only valid for SURE based methods.
Table~\ref{table:methods} summarizes all simulated configurations
including conventional state-of-the-art denoiser, BM3D~\cite{Dabov:2007fh},
that did not require any training, %or the use of 
any ground truth data. Code is available at https://github.com/Shakarim94/Net-SURE.

\begin{table}[!h]
	%\vskip -0.15in
	\caption{Summary of denoising methods. NET can be either SDA or DnCNN.}
	\label{table:methods}
	\centering
	\footnotesize
	%\begin{small}
	%\begin{sc}
	\begin{tabulary}{\columnwidth}{lL}
		\toprule
		Method & Description  \\
		\midrule
		BM3D    		& Conventional state-of-the-art method  \\
		NET-BM3D &  Optimizing MSE with BM3D output as ground truth \\
		%					CNN-SURE-LG    	&   Optimizing SURE with LG \\
		%					CNN-MSE-SM   	&     Optimizing MSE with SM   \\
		NET-SURE    	&  Optimizing SURE without ground truth\\
		%					CNN-SURE-LG-T    &    Optimizing SURE with LG \& T   \\
		%NET-SURE-T 	&  Optimizing SURE without ground truth, but with noisy test data\\
		NET-SURE-FT & Optimizing SURE without ground truth, by fine-tuning on noisy test data\\
		NET-MSE-GT 		&  Optimizing MSE with ground truth \\
		\bottomrule
	\end{tabulary}
	%\end{sc}
	%\end{small}
	%\vskip -0.2in
\end{table}

\subsection{Results:  MNIST dataset}
\label{mnist}
We performed denoising simulations with the MNIST dataset.
Noisy images were generated based on model (\ref{eq:signal}) with two
noise levels (%one with 
$\sigma$ = 25, 50). %and the other with 
%$\sigma$ = 50). 
%
For the experiments on the MNIST dataset which comprised $28 \times 28$ pixels, 
a simple SDA network was chosen~\cite{Vincent:2010vu}.
Decoder and encoder networks each consisted of two convolutional layers (conv) with kernel size $3 \times 3$ and 
sigmoid activation functions, each of which had a stride of two (both conv and conv transposed). 
Thus, a training sample with a size of $28 \times 28$ is downsampled to $7 \times 7$, and then upsampled to $28 \times 28$.

SDA was trained to output a denoised image using a set of 55,000 training and 5,000 validation images. 
The performance of the model was tested with 100 images randomly chosen from the default test set of 10,000 images. 
For all cases, SDA was trained with the Adam optimization algorithm~\cite{Kingma:2015us} with the learning rate of $10^{-3}$ 
for 100 epochs. 
The batch size was set to 200 and bigger batch sizes did not improve the performance. The $\epsilon$ value in (\ref{eq:approx}) was set to $10^{-4}$. 

\begin{table}[!b]
	%\vskip -0.1in
	\caption{Performance of trained DNN denoisers % of trained denoisers 
	for MNIST (performance in dB). Averages of 10 experiments are reported.
		%SURE based SDA denoisers yielded comparable to
		%or better than MSE based SDA denoisers as well as %conventional state-of-the-art denoiser (
		%BM3D.
	}
	\label{table:MNIST}
	\centering
	%	\vskip 0.15in
	\footnotesize
	{\begin{tabular}{c|ccc|c}
			\toprule
			Methods 		& BM3D & SDA-REG & SDA-SURE & SDA-MSE-GT\\
			\midrule
			%$\sigma = 25$ & 27.53 & 25.07 & \bf{27.90} & \bf{28.06} & 27.93 \\
			$\sigma = 25$ & 27.53 & 25.07 & \bf{28.35} & 28.35 \\
			$\sigma = 50$ & 21.82 & 19.85 & \bf{25.23}  & 25.24\\
			\bottomrule
	\end{tabular}}
	%\vskip -0.1in
	
	%	\vskip -0.2in
\end{table}

\begin{figure*}[!h]
	\centering
	\subfloat[Noisy]{\includegraphics[trim=0cm 6cm 6cm 0cm, clip=true,width=0.16\textwidth]{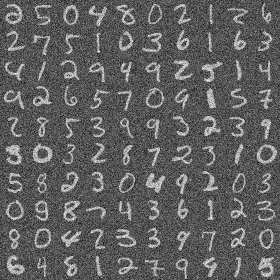}}
	\hfil
	\subfloat[BM3D]{\includegraphics[trim=0cm 6cm 6cm 0cm, clip=true,width=0.16\textwidth]{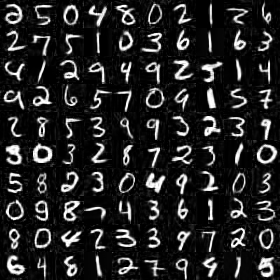}}
	\hfil
	\subfloat[SDA-REG]{\includegraphics[trim=0cm 6cm 6cm 0cm, clip=true,width=0.16\textwidth]{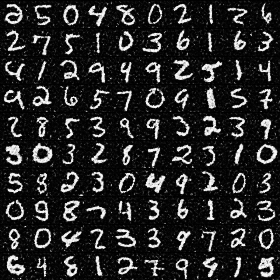}}
	\hfil
	\subfloat[SDA-SURE]{\includegraphics[trim=0cm 6cm 6cm 0cm, clip=true,width=0.16\textwidth]{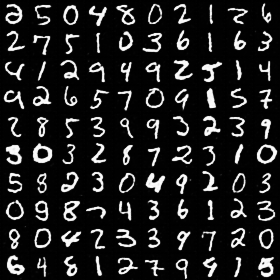}}
	\hfil
	%\subfloat[SDA-SURE-T]{\includegraphics[trim=0cm 6cm 6cm 0cm, clip=true,width=0.16\textwidth]{MNIST/sigma50/AB_sure_50}}
	%\hfil
	\subfloat[SDA-MSE-GT]{\includegraphics[trim=0cm 6cm 6cm 0cm, clip=true,width=0.16\textwidth]{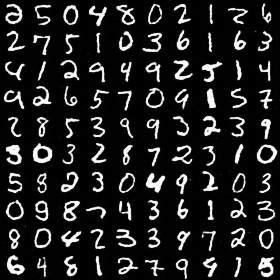}}
	\caption{Denoising results of SDA with various training methods for MNIST dataset at the noise level of $\sigma=50$.} 
	\label{fig-mnist-50}
\end{figure*}

Fig.~\ref{fig-mnist-50} illustrates the visual quality of the outputs of the simulated denoising methods at the noise level of %high noise levels (
$\sigma$ = 50.
%For , deep learning based approaches 
%seem to have an advantage in denoising
%performance over BM3D significantly as shown in  
All SDA-based methods clearly outperform the
conventional BM3D method based on visual inspection since BM3D image looks blurry compared to other SDA-based results.
In contrast, %, while 
it is indistinguishable for the simulation results among SDA-SURE and SDA-MSE-GT methods. 
These observations were confirmed by the quantitative results
as shown in Table~\ref{table:MNIST}. 
Our proposed method SDA-SURE yielded a comparable %almost the same
performance to SDA-MSE-GT (only 0.01 dB difference) and outperformed the
conventional BM3D for all simulated noise levels, $\sigma$ = 25, 50. %, as shown in Table~\ref{table:MNIST}. % shows  that
%SDA methods yielded up to 0.53 dB 
%better PSNR (Peak Signal to Noise Ratio) than
%Slight improvements were obtained by utilizing noisy test images in training (SDA-SURE-T)
%by 0.04 dB better than SDA-MSE-GT and SDA-SURE, at $\sigma$ = 25.
%However, for $\sigma$ = 50, including noisy test data yielded almost the same performance compared to SDA-MSE-GT and SDA-SURE.
%Therefore, it seems that using test data for training SURE based denoiser is helpful 
%to improve denoising performance.

%All SDA-based methods outperformed BM3D significantly, but 
%there were very small differences among all the SDA methods, even when noisy test data were used.

\subsection{Regularization effect of DNN denoisers}

Parametrization of DNNs with different number of parameters and structures may introduce a regularization effect in training 
denoisers. We further investigated this regularization effect by training the SDA to minimize the MSE
%SDA was also trained to minimize the MSE %of the difference 
between the output of SDA and the input noisy image (SDA-REG). 
In the case of the noise level of $\sigma$ = 50, early stopping rule was applied when
%since 
the network started to overfit the noisy dataset after the first few epochs. 
The performance of this method was significantly worse than those of 
all other methods with PSNR values of 25.07 dB ($\sigma$ = 25) and 19.85 dB ($\sigma$ = 50),
as shown in Table~\ref{table:MNIST}. These values are approximately 2 dB lower than the PSNRs of BM3D. 
Fig.~\ref{fig-mnist-50} (c) shows that some noise patterns are still visible %, as shown in . 
even though the regularization effect of SDA greatly reduced noise.
This illustrates that the good performance of SDA is not only attributed to % does not arise just from 
its structure, but %comes from 
also depends on the optimization of MSE or SURE.

\subsection{Accuracy of MC-SURE approximation}

\begin{figure}[!t]
	\begin{minipage}[t]{\linewidth}
		\centering
		{\includegraphics[width=0.85\textwidth]{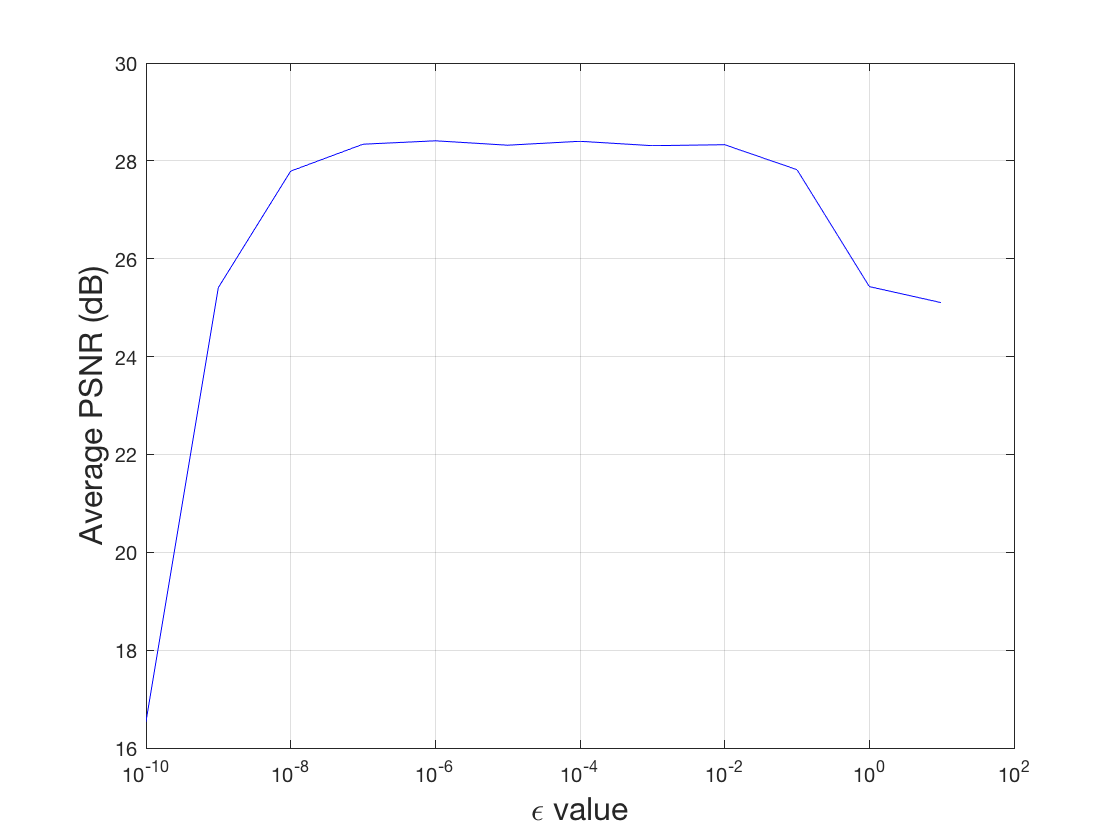}}
		\caption{Performance of SDA-SURE for different $\epsilon$ values at $\sigma= 25$.}
		\label{fig-sens}
	\end{minipage}%
	\hfill
	\begin{minipage}[t]{\linewidth}
		\centering
		{\includegraphics[width=0.85\textwidth]{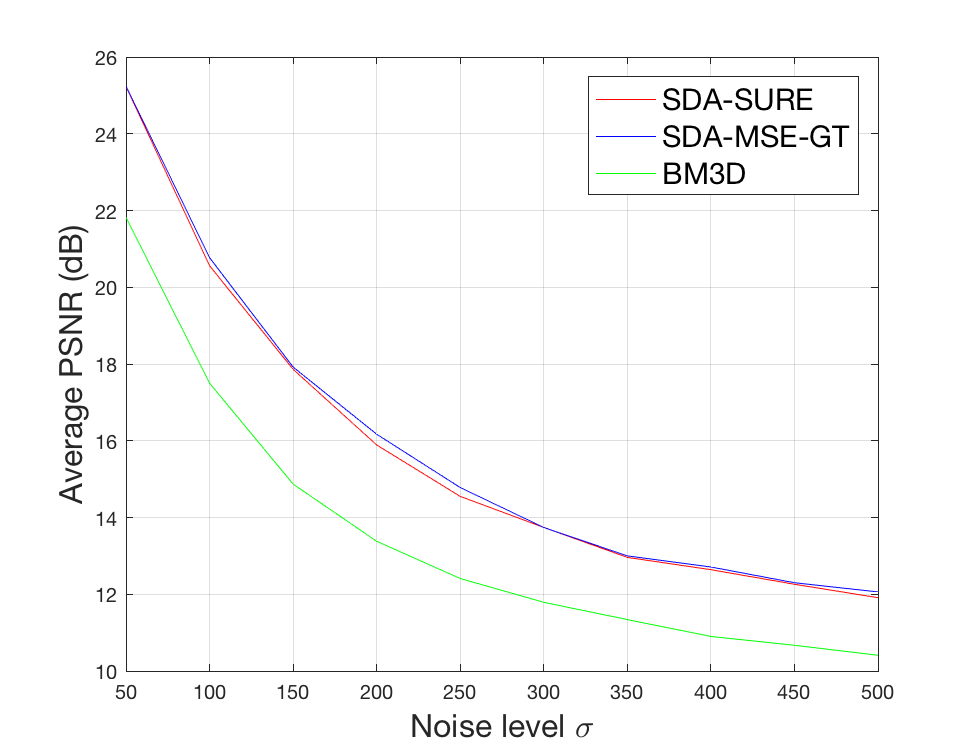}}
		\caption{Performance of denoising methods at different $\sigma$ values.}
		\label{fig-break}
	\end{minipage}
	%\vskip -0.2in
\end{figure}

A small value must be assigned to $\epsilon$ in (\ref{eq:approx}) for accurate estimation of SURE.
Ramani \textit{et al.}~\cite{Ramani:2008ij} have observed that $\epsilon$ can take a wide range of values 
and its choice is not critical.
According to our preliminary experiments for the SDA with an MNIST dataset,
any choice for $\epsilon$ in the range of $[10^{-2}, 10^{-7}]$ worked well so that 
the SURE approximation matches close to 
the MSE during training,
%as illustrated in Figure~\ref{fig-epsilon} (middle).
as illustrated in Fig.~\ref{fig-sens}.
% the curves of true cost function and its SURE approximation follow each other closely during training.
%Extremely small values $\epsilon < 10^{-8}$ resulted in numerical instabilities,
%due to finite machine precision 
%as shown in Figure~\ref{fig-epsilon} (right).
%On the contrary, when $\epsilon > 10^{-1}$, the approximation in (\ref{eq:approx}) becomes substantially inaccurate.
%Figure~\ref{fig-sens} illustrates how the performance of SDA-SURE is affected by the $\epsilon$ value.
%
However, note that these values are only for SDA trained with the MNIST dataset.
The admissible range of $\epsilon$ %, however, 
depends on the DNN denoiser $\boldsymbol{h} ( \boldsymbol{y}; {\boldsymbol \theta} )$.
For example, we observed that 
a suitable $\epsilon$ value must be carefully selected in other cases, such as DnCNN with
large-scale parameters and high resolution images for improved performance.

%\begin{figure*}[!h]
%	\centering
%%	{\includegraphics[width=0.33\textwidth]{epsilon/eps1.png}}
%%	\hfil
%%	{\includegraphics[width=0.33\textwidth]{epsilon/eps1e5.png}}
%%	\hfil
%%	{\includegraphics[width=0.33\textwidth]{epsilon/eps1e9.png}}
%	\caption{Loss curves for the training of SDA with MSE (blue) and its corresponding MC-SURE (red)
%		using different $\epsilon$ values, $\epsilon = 1$ (left), $\epsilon = 10^{-5}$ (middle), and $\epsilon = 10^{-9}$ (right).
%		MC-SURE accurately approximates the true MSE for a wide range of $\epsilon$.
%	}
%	\label{fig-epsilon}
%	%\vskip -0.2in
%\end{figure*}

The accuracy of MC-SURE also depends on the noise level $\sigma$. It was observed that the SURE loss curves become
noisier compared to MSE loss curves as $\sigma$ increases. However, they still followed similar trends and 
yielded similar average PSNRs on MNIST dataset as shown in Fig.~\ref{fig-break}. We observed that after $\sigma=350$,
SURE loss curves started to become too noisy and thus deviated from the trends of their corresponding MSE loss curves. 
Conversely, noise levels $\sigma>300$ were too high for both SDA-based denoisers and BM3D, so that they were not able to output recognizable digits. Therefore, SDA-SURE can be trained effectively on adequately high noise levels 
so that it can yield a performance that is
comparable to SDA-MSE-GT and can consistently outperform BM3D.

\begin{table*}[!t]
	%	\vskip -0.1in
	\caption{Results of denoising methods on 12 widely used images (Set12 dataset) (Performance in dB).}
	\label{table:DeepCNN}
	%	\vskip 0.15in
	\centering
	%\resizebox{\textwidth}{!}
	{\begin{tabulary}{\textwidth}{l | CCcCCCcCcCCc | c}
			\midrule
			Image & C.Man & House & Peppers & Starfish & Monar. & Airpl. & Parrot & Lena & Barbara & Boat & Man & Couple & \bf{Average} \\
			\midrule
			\multicolumn{14}{c}{$\sigma = 25$}\\
			\midrule
			BM3D    		& 29.47 & \bf{33.00} & 30.23 & 28.58 & 29.35	& 28.37	& 28.89 & \bf{32.06} & \bf{30.64} & 29.78 & 29.60 & 29.70 & 29.97\\
			DnCNN-BM3D & 29.34 & 31.99 & 30.13 & 28.38 & 29.21 & 28.46 & 28.91 & 31.53 & 28.89 & 29.6 & 29.52 & 29.54 & 29.63\\
			DnCNN-SURE	& \bf{29.80} & 32.70 & \bf{30.58} & \bf{29.08} & \bf{30.11} & \bf{28.94} & \bf{29.17} & \bf{32.06} & 29.16 & \bf{29.84} & \bf{29.89} & \bf{29.76} & \bf{30.09}\\
			%\midrule
			%DnCNN-SURE-T & \bf{29.86} & 32.73 & 30.57 & \bf{29.11} & \bf{30.13} & 28.93 & \bf{29.26} & \bf{32.08} & 29.44 & \bf{29.86} & \bf{29.91} & \bf{29.78} & \bf{30.14}\\
			\midrule
			DnCNN-MSE-GT & 30.14 & 33.16 & 30.84 & 29.4 & 30.45 & 29.11 & 29.36 & 32.44 & 29.91 & 30.11 & 30.08 & 30.06 & 30.42\\
			\midrule
			\multicolumn{14}{c}{$\sigma = 50$}\\
			\midrule
			BM3D    		& 26.00 & \bf{29.51} & 26.58 & 25.01 & 25.78 & 25.15 & 25.98 & \bf{28.93} & \bf{27.19} & 26.62 & 26.79 & 26.46 & \bf{26.67} \\
			DnCNN-BM3D & 25.76 & 28.43 & 26.5 & 24.9 & 25.66 & 25.15 & 25.82 & 28.36 & 25.3 & 26.5 & 26.6 & 26.17 & 26.26 \\
			DnCNN-SURE	& \bf{26.48} & 29.14 & \bf{26.77} & \bf{25.38} & \bf{26.50} & \bf{25.66} & \bf{26.21} & 28.79 & 24.86 & \bf{26.78} & \bf{26.97} & \bf{26.51} & \bf{26.67}\\
			%\midrule
			%DnCNN-SURE-T & 26.47 & 29.20 & \bf{26.78} & \bf{25.39} & \bf{26.53} & 25.65 & \bf{26.21} & 28.81 & 25.23 & \bf{26.79} & \bf{26.97} & 26.48 & \bf{26.71}\\
			\midrule
			DnCNN-MSE-GT & 27.03 & 29.92 & 27.27 & 25.65 & 26.95 & 25.93 & 26.43 & 29.31 & 26.17 & 27.12 & 27.22 & 26.94 & 27.16\\
			\midrule
			\multicolumn{14}{c}{$\sigma = 75$}\\
			\midrule
			BM3D    		& 24.58 & \bf{27.45} & \bf{24.69} & 23.19 & 23.81 & 23.38 & \bf{24.22 }& \bf{27.14} & \bf{25.08} & 25.05 & 25.30 & \bf{24.73} & \bf{24.89}\\
			DnCNN-BM3D & 24.11 & 27.02 & 24.48 & 23.09 & 23.73 & 23.40 & 24.06 & 27.11 & 23.80 & 24.84 & 25.19 & 24.59 & 24.62\\
			DnCNN-SURE	& \bf{24.65} & 27.16 & 24.49 & \bf{23.25} & \bf{24.10} & \bf{23.52} & 24.13 & 26.92 & 23.02 & \bf{25.09} & \bf{25.37} & 24.70 & 24.70\\
			%\midrule
			%DnCNN-SURE-T & \bf{24.82} & 27.34 & 24.58 & \bf{23.34} & \bf{24.25} & \bf{23.56} & \bf{24.44} & 27.03 & 23.07 & \bf{25.17} & \bf{25.45} & \bf{24.78} & 24.82\\
			\midrule
			DnCNN-MSE-GT   & 25.46 & 28.04 & 25.22 & 23.62 & 24.81 & 23.97 & 24.71 & 27.60 & 23.88 & 25.53 & 25.68 & 25.13 & 25.30\\
			\bottomrule
		\end{tabulary}}

	%	\vskip -0.1in
\end{table*}

\subsection{Results: high resolution natural images dataset} % with Deep CNN Image Denoiser}
\label{DnCNN}
To demonstrate the capabilities of our SURE-based deep learning denoisers, 
we investigated %experimented on denoising 
%larger images which require training of 
a deeper and more powerful denoising network called DnCNN~\cite{Zhang:2017eh}
using high resolution images. 
%For this task we used 
%DnCNN-S network~\cite{Zhang:2017eh}
%. proposed by Zhang et al., proposed 
DnCNN consisted of %using a 
17 layers of CNN with batch normalization and ReLU activation functions.
% as shown in Figure~\ref{fig-DnCNN}.
Each convolutional layer had 64 filters with sizes of $3\times3$. 
Similar to~\cite{Zhang:2017eh}, %we train 
the network was trained with 400 images with matrix sizes of $180 \times 180$ pixels. 
In total, $1772 \times 128$ image patches 
with sizes of $40 \times 40$ pixels were extracted randomly 
from these images. 
%As in~\cite{Zhang:2017eh}, 
Two test sets were used to evaluate performance: one set consisted of 12 widely used images (Set12)~\cite{Dabov:2007fh}, and the other was a BSD68 dataset.
%For DnCNN-SURE-T, additional $808 \times 128$ image patches 
%were extracted from these noisy test images, and were then added to the training dataset.
For all cases, the network was trained with 50 epochs 
using the Adam optimization algorithm with an initial learning rate of $10^{-3}$, which eventually
%and then
decayed to $10^{-4}$ after 40 epochs. The batch size was set to 128 (note that
bigger batch sizes did not improve performance).
Images were corrupted at three noise levels ($\sigma$ = 25, 50, 75).

DnCNN used residual learning whereby the network was forced to learn 
the difference between noisy and ground truth images. % as shown in Figure~\ref{fig-DnCNN}. 
The output residual image was then subtracted 
from the input noisy image to yield %get 
the %actual 
estimated image. %In other words, 
%we can easily train 
Thus, our network was trained with SURE as %below: using (\ref{eq:residual}) 
\begin{equation}
\boldsymbol{h}( \boldsymbol{y}; {\boldsymbol \theta} ) = \boldsymbol{y} - \boldsymbol{\mathrm{CNN}}_{{\boldsymbol \theta}}( \boldsymbol{y} )
\label{eq:residual}
\end{equation}
where $\boldsymbol{\mathrm{CNN}}_{{\boldsymbol \theta}}(.)$ is the DnCNN 
that is being trained using residual learning.
For DnCNN, selecting an appropriate $\epsilon$ value in (\ref{eq:approx}) turned out to be important
for a good denoising performance. To achieve stable training with good performance, $\epsilon$ had to be tuned for each of the chosen noise levels of $\sigma$ = 25, 50, 75. We observed that the optimal value for $\epsilon$ was proportional to $\sigma$ as shown in~\cite{deledalle2014stein}. All the experiments were performed with the setting of 
\begin{equation}
\epsilon = \sigma\times1.4 \times 10^{-4}.
\end{equation}
With the use of an NVidia Titan X GPU, the training process took approximately 7 hours for 
DnCNN-MSE-GT and approximately 11 hours for DnCNN-SURE. 
SURE based methods took more training time than MSE based methods 
because of the additional divergence %term 
calculations executed to optimize the MC-SURE cost function.
%For the DnCNN-SURE-T method, it took approximately 15 hours to complete the training owing to the larger dataset.

\begin{table}[!t]
	%	\vskip -0.1in
	\caption{Results of denoising methods on BSD68 dataset (Performance in dB).}
	\label{table:BSD68}
	%	\vskip 0.15in
	\centering
	\scriptsize
	{\begin{tabulary}{\columnwidth}{c|CCC|C}
			\toprule
			Methods 		& BM3D &DnCNN-BM3D  & DnCNN-SURE & DnCNN-MSE-GT\\
			\midrule
			$\sigma = 25$ & 28.56 & 28.54 & \bf{28.97}  & 29.20 \\
			$\sigma = 50$ & 25.62 & 25.44 & \bf{25.93}  & 26.22\\
			$\sigma = 75$ & 24.20 & 24.09 & \bf{24.31}  & 24.66\\
			\bottomrule
	\end{tabulary}}
	%	\vskip -0.1in
\end{table}

Tables~\ref{table:DeepCNN} and~\ref{table:BSD68} present denoising performance results using (a) the BM3D denoiser~\cite{Dabov:2007fh}, 
(b) a state-of-the-art deep CNN (DnCNN) image denoiser trained with MSE~\cite{Zhang:2017eh}, and
(c) the same DnCNN image denoiser trained with SURE without the use of noiseless ground truth images.
%for different dataset variations (as shown in Table~\ref{table:methods}).
The MSE-based DnCNN image denoiser with ground truth data, DnCNN-MSE-GT, 
yielded the best denoising performance compared to other
methods, such as the BM3D, which is consistent with the results in~\cite{Zhang:2017eh}.

As seen in Table~\ref{table:DeepCNN}, for the Set12 dataset, SURE-based denoisers achieved performances comparable to or better than that for BM3D for noise levels $\sigma = 25$ and $50$.
In contrast, for a higher noise level ($\sigma = 75$), DnCNN-SURE %and DnCNN-SURE-T 
yielded 
lower average PSNR value by 0.19 dB than BM3D. 
%DnCNN-SURE-T outperformed DnCNN-SURE in all cases, and had considerably better performance on some images, such as ``Barbara.''
BM3D had exceptionally good denoising performance on the ``Barbara'' image (up to 2.33 dB better PSNR), and even outperformed the DnCNN-MSE-GT method.
In the case of the BSD68 dataset in Table~\ref{table:BSD68}, the SURE-based method outperformed BM3D for all the noise levels.
Unlike the case of Set12, we observed that DnCNN-SURE yielded significantly better performance than BM3D, and yielded increased average PSNR values by 0.11-0.41 dB. %higher average PSNR.
%It was also observed that DnCNN-SURE-T benefited from the utilization of noisy test images  and %so that it further 
%improved the average PSNR of DnCNN-SURE.

Differences among the performances of denoisers in Tables~\ref{table:DeepCNN} and~\ref{table:BSD68} can be explained by the working principle of BM3D.
Since BM3D looks for similar image patches for denoising, repeated patterns (as in the ``Barbara'' image) and flat areas (as in ``House'' image) can be key factors to generating %yield strong 
improved denoising results. One of the advantages of DnCNN-SURE over BM3D is that it does not suffer from rare patch effects. If the test image is relatively detailed and does not contain many repeated patterns, BM3D will have poorer performance than the proposed DnCNN-SURE method.
Note that the DnCNN-BM3D method that trains networks by optimizing MSE with BM3D denoised images as the ground truth yielded slightly worse performance than the BM3D itself 
%data and DnCNN was trained by optimizing MSE. This method also did not require noiseless images, however had the worst performance among all denoising methods including BM3D as shown in
(Tables~\ref{table:DeepCNN},~\ref{table:BSD68}).

\begin{figure*}[!t]
	%\vskip 0.2in
	%\captionsetup{textfont=normalfont}
	%\captionsetup[sub]{textfont=scriptsize}
	\centering
	\subfloat[Noisy image / 14.76dB]{\includegraphics[width=0.24\textwidth]{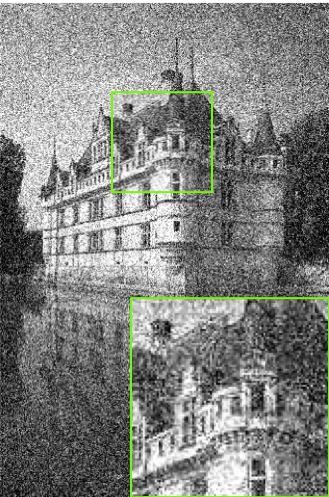}}
	\hfil
	\subfloat[BM3D / 26.14dB]{\includegraphics[width=0.24\textwidth]{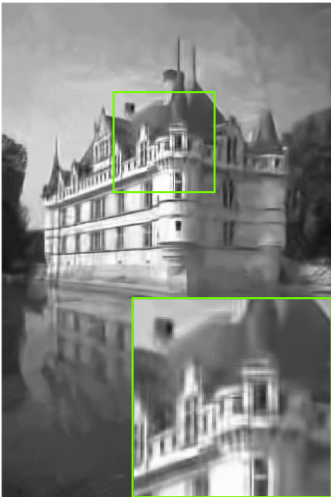}}
	\hfil
	\subfloat[DnCNN-SURE / 26.46dB]{\includegraphics[width=0.24\textwidth]{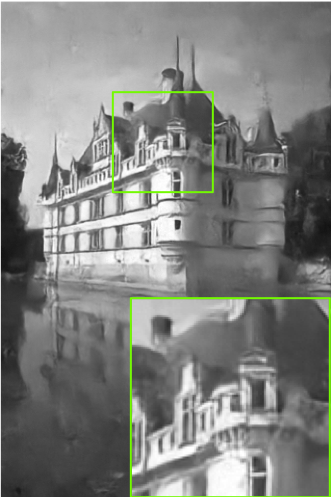}}
	%\hfil
	%\subfloat[SURE-T / 26.46dB]{\includegraphics[width=0.19\textwidth]{DnCNN/15_SURE_T_sigma50.png}}
	\hfil
	\subfloat[DnCNN-MSE-GT / 26.85dB]{\includegraphics[width=0.24\textwidth]{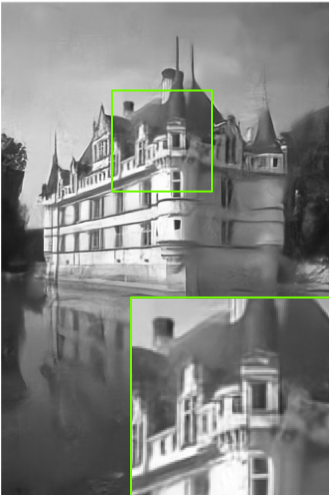}}	
	\caption{Denoising results of an image from the BSD68 dataset for $\sigma=50$.}
	\label{fig-dncnn}
	%	\vskip -0.2in
\end{figure*}
\begin{figure*}[!t]
	%\vskip 0.2in
	%\captionsetup{textfont=normalfont}
	%\captionsetup[sub]{textfont=scriptsize}
	\centering
	\subfloat[Noisy image / 11.77dB]{\includegraphics[width=0.24\textwidth]{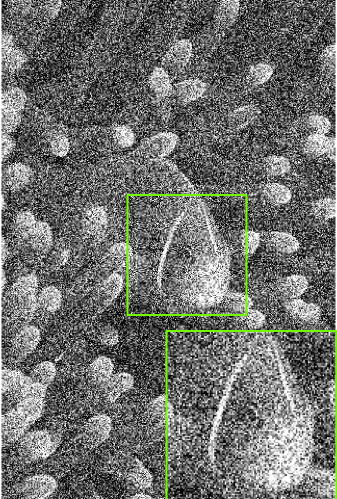}}
	\hfil
	\subfloat[BM3D / 26.21dB]{\includegraphics[width=0.24\textwidth]{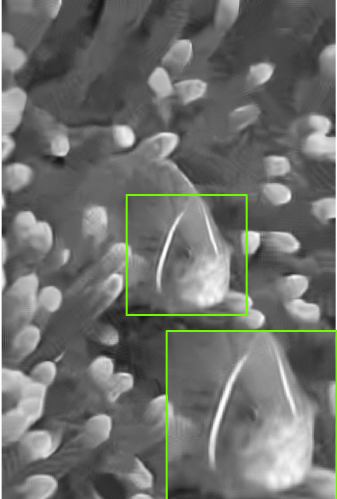}}
	\hfil
	\subfloat[DnCNN-SURE / 26.46dB]{\includegraphics[width=0.24\textwidth]{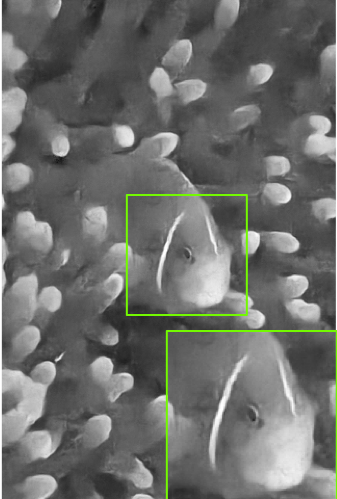}}
	%\hfil
	%\subfloat[SURE-T / 26.67dB]{\includegraphics[width=0.19\textwidth]{DnCNN/21_SURE_T_sigma75.png}}
	\hfil
	\subfloat[DnCNN-MSE-GT / 27.10dB]{\includegraphics[width=0.24\textwidth]{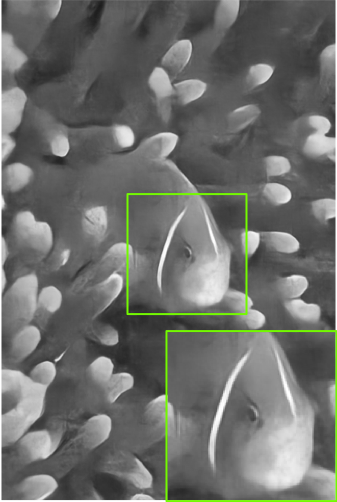}}
	\caption{Denoising results of an image from the BSD68 dataset for $\sigma=75$.}
	\label{fig-dncnn2}
	%	\vskip -0.2in
\end{figure*}

Fig.~\ref{fig-dncnn} illustrates the denoised results for an image from the BSD68 dataset. Visual quality assessment indicated that 
BM3D yielded blurrier images
and thus yielded worse PSNR compared to the results generated by DNN denoisers.
DnCNN-SURE effectively removed noise while preserving edges to %and 
yield sharper images.
DnCNN-MSE-GT had the best denoised image with the highest PSNR of 26.85 dB.
In Fig.~\ref{fig-dncnn2}, an image from the BSD68 dataset is contaminated with severe noise ($\sigma=75$).
BM3D struggled to preserve important details in the image such as the outline of the fish's eye, while DnCNN-SURE
yielded much sharper image with higher PSNR.
%while both SURE methods yielded very similar performances in accordance with PSNR and visual quality assessment.

\subsection{Extension to blind color image denoising}
\label{CDnCNN}

We adopted the CDnCNN network from~\cite{Zhang:2017eh} for blind color denoising  and incorporated our
SURE-based training by minimizing (\ref{eq:riskdenoiser_emp_batch_prop_mcsure_multi}). 
Following~\cite{Zhang:2017eh}, the network consisted of 20 layers and was trained with 432 colored images.
In total, $3000\times128$ image patches with sizes of $50\times50$ were randomly extracted from these images.
We trained a single CDnCNN network for the noise levels %in the range 
of $\sigma=\left[0, 55\right]$.
The experiments were performed setting $\epsilon^{(j)} = \sigma^{(j)} \times 1.2 \times 10^{-4}$. 
All the other simulation details were the same as the ones for grayscale denoising in Section~\ref{DnCNN}.

\begin{table}[!b]
	%	\vskip -0.1in
	\caption{Results of denoising methods on CBSD68 dataset (Performance in dB).}
	\label{table:CBSD68}
	%	\vskip 0.15in
	\centering
	\scriptsize
	{\begin{tabular}{c|cc|c}
			\toprule
			Methods 		& CBM3D &CDnCNN-SURE  & CDnCNN-MSE-GT\\
			\midrule
			$\sigma = 25$ & 30.70 & \bf{31.06} & 31.21 \\
			$\sigma = 50$ & 27.38 & \bf{27.75} & 27.96\\
			\bottomrule
	\end{tabular}}
	%	\vskip -0.1in
\end{table}

\begin{figure*}[!t]
	%\vskip 0.2in
	%\captionsetup{textfont=normalfont}
	%\captionsetup[sub]{textfont=scriptsize}
	\centering
	\subfloat[Noisy image / 14.67dB]{\includegraphics[width=0.24\textwidth]{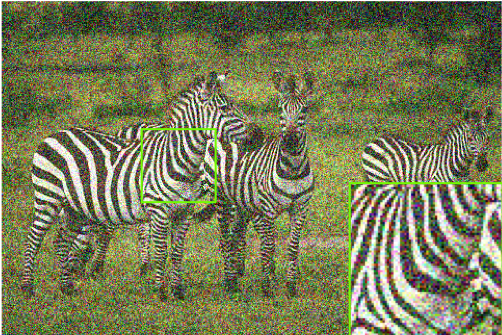}}
	\hfil
	\subfloat[CBM3D / 26.47dB]{\includegraphics[width=0.24\textwidth]{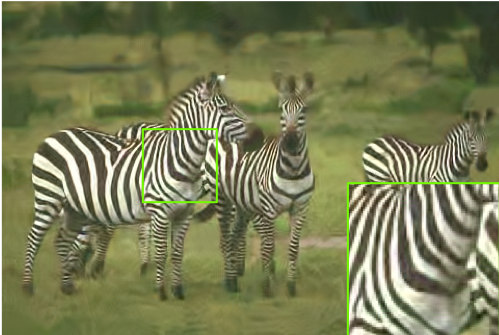}}
	\hfil
	\subfloat[CDnCNN-SURE / 27.28dB]{\includegraphics[width=0.24\textwidth]{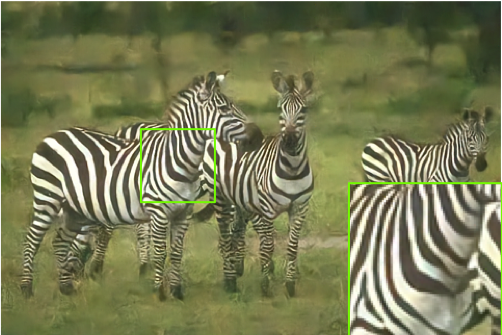}}
	\hfil
	\subfloat[CDnCNN-MSE-GT / 27.47dB]{\includegraphics[width=0.24\textwidth]{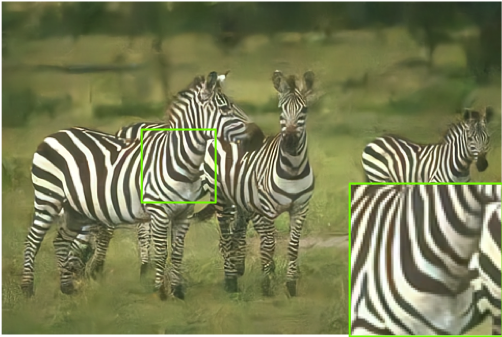}}
	
	\caption{Denoising results of an image from the CBSD68 dataset for $\sigma=50$.}
	\label{fig-cdncnn}
	%	\vskip -0.2in
\end{figure*}

\begin{table*}[!h]
	%	\vskip -0.1in
	\caption{Results of denoising methods on 9 widely used color images (Performance in dB). \textbf{*} 
	indicates instances where SURE-FT outperformed MSE-GT.}
	\label{table:blindcolor}
	%	\vskip 0.15in
	\centering
	%\resizebox{\textwidth}{!}
	{\begin{tabular}{l | ccccccccc | c}
			\midrule
			Image & House & Peppers & Lena & Baboon & F16 & Kodak1 & Kodak2 & Kodak3 & Kodak12  & \bf{Average} \\
			\midrule
			\multicolumn{11}{c}{$\sigma = 25$}\\
			\midrule
			CBM3D    		& \bf{33.02} & 31.23 & 32.27 & 25.95 & 32.76 & 29.17 & 32.44 & 34.61 & 33.76 &  31.69\\
			DIP    		& 31.75 & 29.09 & 30.50 & 22.67 & 31.56 & 25.21 & 30.39 & 30.56 & 30.71 &  29.16\\
			CDnCNN-SURE & 31.06 & 30.35 & 31.88 & 25.53 & 31.68 & 29.54 & 32.56 & 34.42 & 33.70 & 31.19 \\
			CDnCNN-SURE-FT	& 32.69* & \bf{31.35*} & \bf{32.45*} & \bf{26.47*} & \bf{33.30*} & \bf{29.77*} & \bf{32.90*} & \bf{34.72} & \bf{33.94} & \bf{31.95*}\\
			\midrule
			CDnCNN-MSE-GT & 31.47 & 30.67 & 32.13 & 25.82 & 31.95 & 29.64 & 32.83 & 34.77 & 33.98 & 31.47\\
			\midrule
			\multicolumn{11}{c}{$\sigma = 50$}\\
			\midrule
			CBM3D    		& \bf{30.54} & 28.97 & \bf{29.90} & 23.14 & 29.76 & 25.90 & 29.89 & \bf{31.43} & \bf{30.98} & 28.95 \\
			DIP    		& 28.39 & 26.98 & 28.41 & 21.21 & 27.79 & 23.96 & 27.52 & 28.09 & 27.63 &  26.66\\
			CDnCNN-SURE & 29.01 & 28.29 & 29.44 & 23.21 & 28.99 & 26.29 & 29.76 & 31.19 & 30.79 & 28.55 \\
			CDnCNN-SURE-FT	& 29.95* & \bf{29.03*} & 29.86* & \bf{23.64*} & \bf{30.14*} & \bf{26.50*} & \bf{30.09} & \bf{31.43} & 30.92 & \bf{29.06*}\\
			\midrule
			CDnCNN-MSE-GT & 29.57 & 28.69 & 29.74 & 23.40 & 29.36 & 26.45 & 30.21 & 31.66 & 31.19 & 28.92\\
			\bottomrule
	\end{tabular}}
	%	\vskip -0.1in
\end{table*}

Table \ref{table:CBSD68} presents denoising performance on CBSD68 dataset
using  (a) the CBM3D denoiser~\cite{Dabov:2007fh},
(b) CDnCNN image denoiser trained with MSE~\cite{Zhang:2017eh}, and
(c) the same CDnCNN image denoiser trained with SURE without the use of noiseless ground truth images.
Both deep learning methods outperformed the conventional method CBM3D method by a large margin. 
This is consistent with the results of DnCNN methods on the grayscale BSD68 dataset.
Even though color blind image denoising is a harder task than grayscale non-blind single noise level denoising,
CDnCNN-SURE showed a superior denoising performance compared to the CBM3D. 
This illustrates the powerful capabilities of the SURE-based optimization for extended settings such as multiple noise levels and color images.

Fig.~\ref{fig-cdncnn} illustrates the denoised results for an image from CBSD68 dataset that was contaminated 
at the noise level of $\sigma=50$. 
Both deep learning methods yielded higher quality images compared to the CBM3D. 
The denoised image CBM3D contained false color artifacts whereas CDnCNN methods preserved the true colors and fine details.

\subsection{Unsupervised refining (fine-tuning) with SURE}
\label{SURE-FT}

Since our SURE-based training method does not require the ground truth images, we can utilize noisy test images
to train the deep neural networks. One way to execute this could be just adding the noisy test images to the 
training datasets and train the network from scratch~\cite{NIPS2018_7587}. 
However, this method is slow and requires retraining the network for the future test sets.
We propose a more practical and faster method called SURE-FT in which we fine-tune 
a pretrained denoising network on a noisy test image by minimizing (\ref{eq:riskdenoiser_refine}).

To demonstrate the effectiveness of this method, we took CDnCNN-SURE from Section \ref{CDnCNN} as a baseline denoiser network and fine-tuned it on an individual noisy  image (CDnCNN-SURE-FT) for all test images. Our proposed fine-tuning process
is different from %had slight differences from 
the original training process. Firstly, we used a single test image without dividing it into 
small patches for each fine-tuning. Secondly, we froze batch normalization layers in CDnCNN because of the change of the dataset size.
The initial learning rate was set to $10^{-4}$ and the network was fine-tuned for 75 epochs 
(learning rate was decayed to $5 \times 10^{-5}$ after 50 epochs). For each image, it took about 24.5, 75.25, 123.75 seconds per image (256 $\times$ 256, 512 $\times$ 512, , 768 $\times$ 512), respectively, for 75 epochs.

The methods are evaluated on 9 widely used color images~\cite{foiwebsite} and CDnCNN-SURE-FT was implemented
for each image separately. We additionally report the performance of deep image prior method (DIP)~\cite{ulyanov2017deep}, 
which also optimizes DNNs using the input noisy image only. 
We used the official PyTorch implementation %({https://github.com/DmitryUlyanov/deep-image-prior}) and
from the authors and took the hyperparameters from the paper~\cite{ulyanov2017deep}.

Table \ref{table:blindcolor} demonstrates the performance of various denoising methods. Unlike the CBSD68 dataset,
CBM3D has superior performance to both CDnCNN-SURE and CDnCNN-MSE-GT methods on most of the color images. %the 9 color images.
The structure of these test images are quite different from the 432 training images and some contain many repeated
patterns and flat areas which are favorable for CBM3D. This is where fine-tuning on the noisy test images can provide
its benefits. %As we can see the 
Our proposed CDnCNN-SURE-FT vastly improved the performance over the CDnCNN-SURE on almost all of the test images, often by more than 1dB gain in some images.
This shows that the network could learn the unique patterns and details from the noisy test images 
and denoised them effectively. As a result, CDnCNN-SURE-FT outperformed all the other methods on both noise levels, including CBM3D, CDnCNN-MSE-GT in many images and on average (indicated with *).
The DIP method had the worst performance, falling behind the CDnCNN-SURE method by almost 2dB.

\begin{figure*}[!h]
	%\vskip 0.2in
	%\captionsetup{textfont=normalfont}
	%\captionsetup[sub]{textfont=scriptsize}
	\centering
	\subfloat[Noisy image / 20.34dB]{\includegraphics[width=0.32\textwidth]{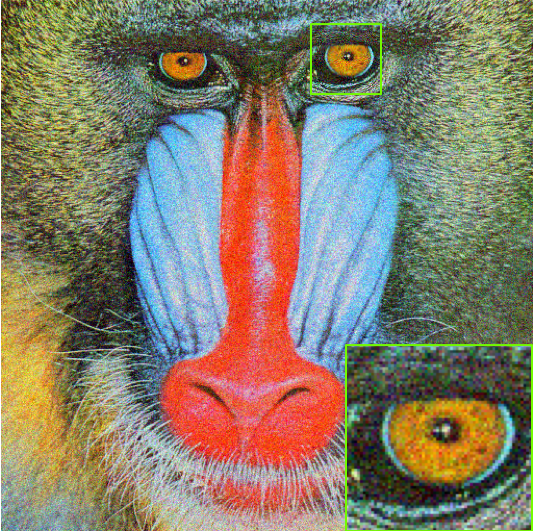}}
	\hfil
	\subfloat[Ground truth ]{\includegraphics[width=0.32\textwidth]{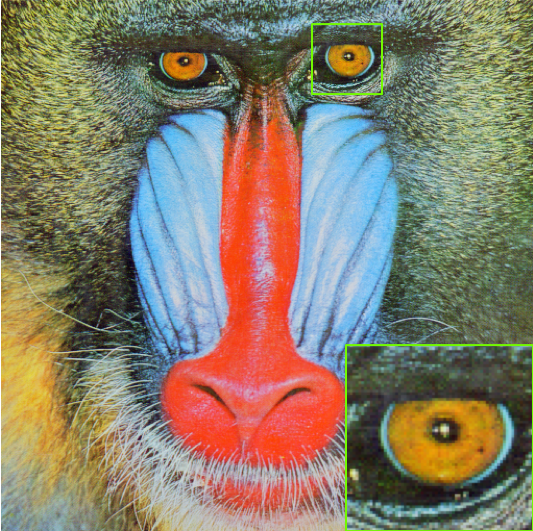}}
	\hfil
	\subfloat[CBM3D / 29.89dB]{\includegraphics[width=0.32\textwidth]{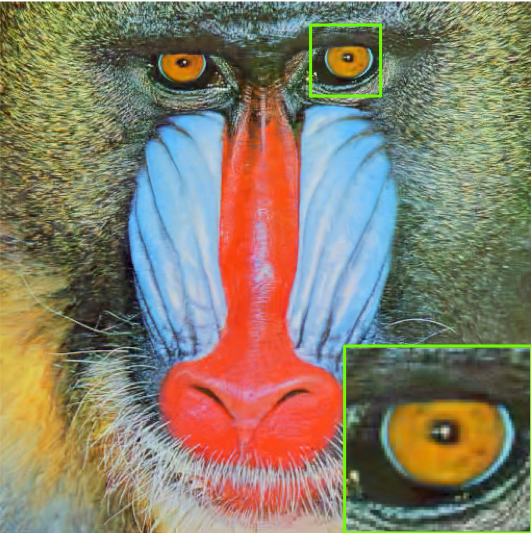}}
	\hfil
	\subfloat[DIP / 27.52dB]{\includegraphics[width=0.32\textwidth]{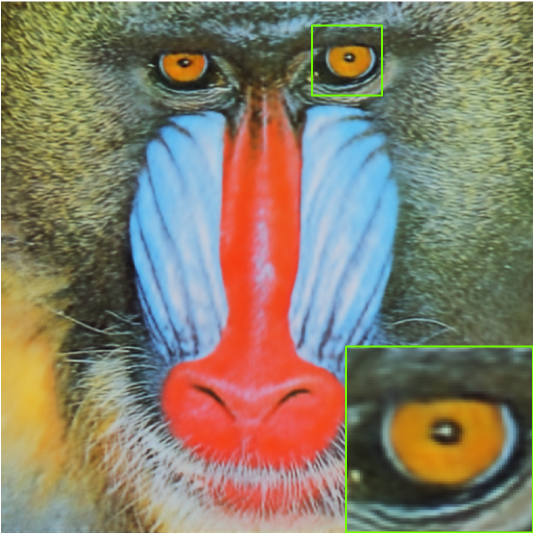}}
	\hfil
	\subfloat[CDnCNN-SURE / 29.76dB]{\includegraphics[width=0.32\textwidth]{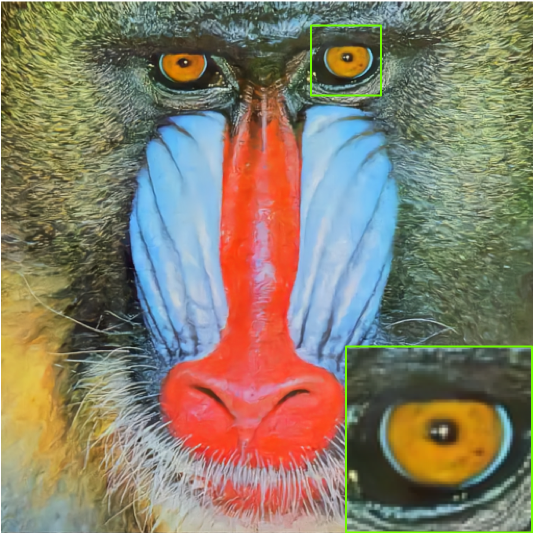}}
	\hfil
	\subfloat[CDnCNN-SURE-FT / 30.09dB]{\includegraphics[width=0.32\textwidth]{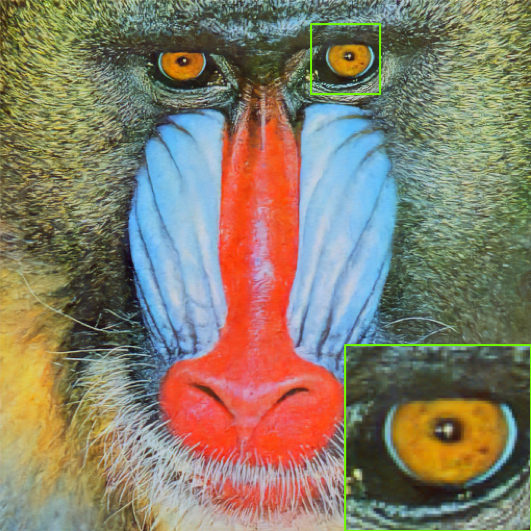}}	
	\caption{Denoising results of the ``Baboon" image for $\sigma=25$.}
	\label{fig-baboon}
	%	\vskip -0.2in
\end{figure*}
\begin{figure*}[!h]
	%\vskip 0.2in
	%\captionsetup{textfont=normalfont}
	%\captionsetup[sub]{textfont=scriptsize}
	\centering
	\subfloat[Noisy image / 15.21dB ]{\includegraphics[width=0.32\textwidth]{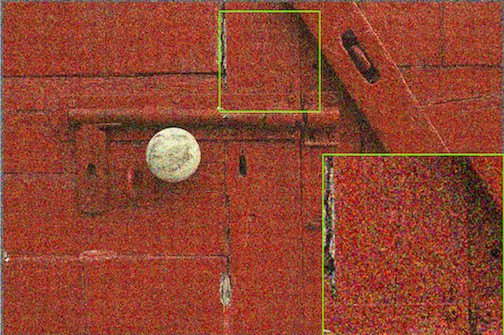}}
	\hfil
	\subfloat[Ground Truth ]{\includegraphics[width=0.32\textwidth]{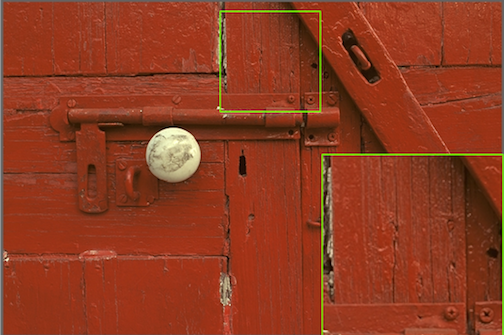}}
	\hfil
	\subfloat[CBM3D / 32.27dB]{\includegraphics[width=0.32\textwidth]{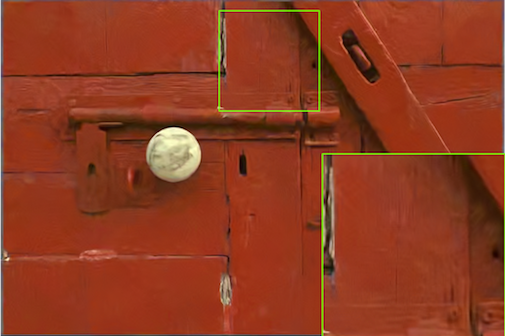}}
	\hfil
	\subfloat[DIP / 30.50dB]{\includegraphics[width=0.32\textwidth]{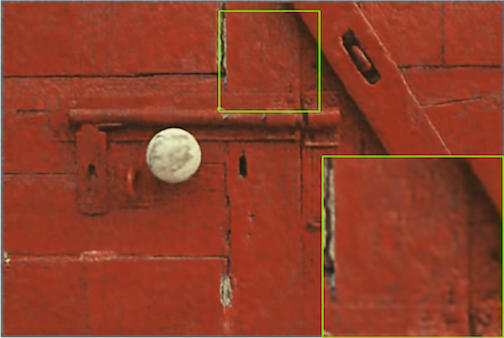}}
	\hfil
	\subfloat[CDnCNN-SURE / 31.88dB]{\includegraphics[width=0.32\textwidth]{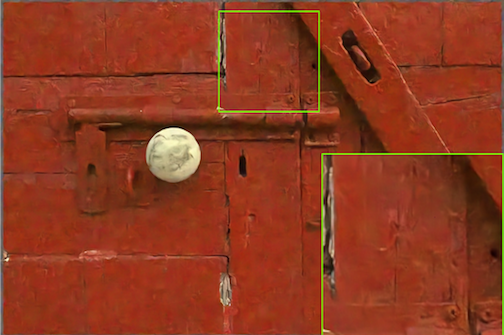}}
	\hfil
	\subfloat[CDnCNN-SURE-FT / 32.45dB]{\includegraphics[width=0.32\textwidth]{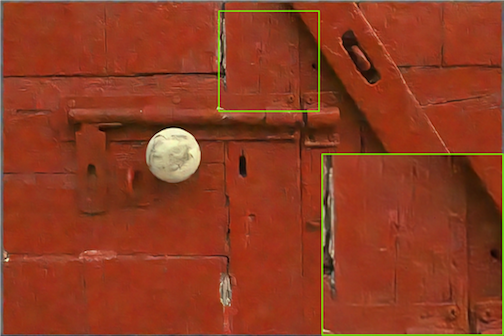}}
	\caption{Denoising results of the ``Kodak 2" image for $\sigma=50$.}
	\label{fig-kodak2}
	%	\vskip -0.2in
\end{figure*}

\begin{figure*}[!h]
	\centering
	\subfloat[Noisy]{\includegraphics[trim=0cm 6cm 6cm 0cm, clip=true,width=0.16\textwidth]{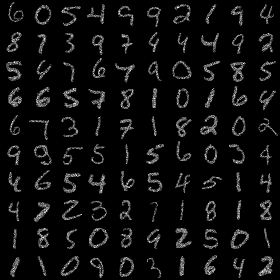}}
	\hfil
	\subfloat[BM3D+VST]{\includegraphics[trim=0cm 6cm 6cm 0cm, clip=true,width=0.16\textwidth]{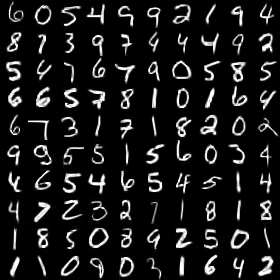}}
	\hfil
	\subfloat[SDA-PURE]{\includegraphics[trim=0cm 6cm 6cm 0cm, clip=true,width=0.16\textwidth]{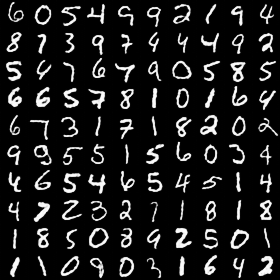}}
	\hfil
	\subfloat[SDA-MSE-GT]{\includegraphics[trim=0cm 6cm 6cm 0cm, clip=true,width=0.16\textwidth]{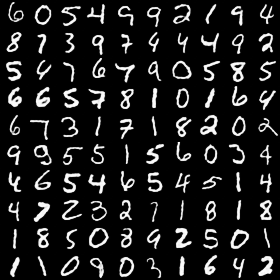}}
	\caption{Denoising results of SDA with various methods for MNIST dataset for $\zeta=0.1$.} 
	\label{fig-mnist-pure}
\end{figure*}
\begin{figure*}[!h]
	\centering
	\subfloat[Noisy]{\includegraphics[trim=6cm 0cm 0cm 6cm, clip=true,width=0.16\textwidth]{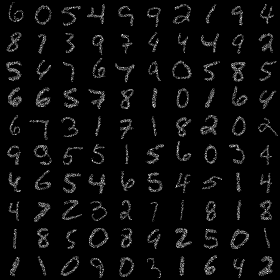}}
	\hfil
	\subfloat[BM3D+VST]{\includegraphics[trim=6cm 0cm 0cm 6cm, clip=true,width=0.16\textwidth]{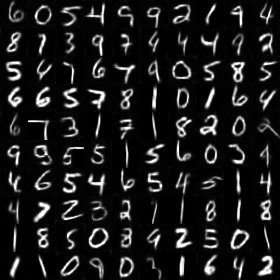}}
	\hfil
	\subfloat[SDA-PURE]{\includegraphics[trim=6cm 0cm 0cm 6cm, clip=true,width=0.16\textwidth]{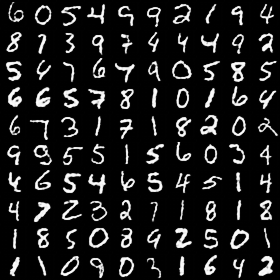}}
	\hfil
	\subfloat[SDA-MSE-GT]{\includegraphics[trim=6cm 0cm 0cm 6cm, clip=true,width=0.16\textwidth]{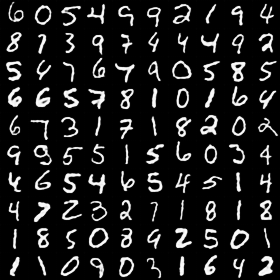}}
	\caption{Denoising results of SDA with various methods for MNIST dataset for $\zeta=0.2$.} 
	\label{fig-mnist-pure2}
\end{figure*}

Fig.~\ref{fig-baboon} illustrates the denoised results of the ``Baboon'' image at $\sigma=25$. 
The eye region is good for the comparison of the methods. We can see that CBM3D had slightly better details and quantitative performance than CDnCNN-SURE, 
while DIP image was relatively blurry. Our proposed CDnCNN-SURE-FT yielded a sharper image which was closer to the ground truth image visually and quantitatively.
In Fig.~\ref{fig-kodak2}, all methods were evaluated on the ``Kodak 2'' image that was contaminated with the noise level of
$\sigma=50$. 
CDnCNN-SURE image seemed to preserve more details than CBM3D (bottom right), however contained artifacts
throughout the entire image. Our CDnCNN-SURE-FT got rid of most artifacts and yielded cleaner image. Moreover,
the denoised image contained more details than CBM3D and was closer to the ground truth. The DIP yielded
blurry and low quality image.

\subsection{Extension to Poisson noise denoising}
\label{PURE}

We performed Poisson denoising simulations with the MNIST dataset with the same SDA network as in Section \ref{mnist}.
For the comparison, we chose the conventional BM3D+VST method~\cite{makitalo2011optimal} that uses the optimal
Anscombe transformation. Table \ref{table:MNIST_pure} shows that for $\zeta=0.01$, all methods had similar performance
in terms of average PSNR. However, for higher $\zeta=0.1, 0.2$ values (higher noise), our proposed SDA-PURE method outperformed the 
conventional BM3D+VST method significantly, while SDA-MSE-GT that was trained with the ground truth data yielded the best results.
Figs. \ref{fig-mnist-pure} and \ref{fig-mnist-pure2} illustrate the visual quality of the outputs of the
denoising methods for $\zeta=0.1$ and $0.2$, respectively. We observed that BM3D+VST images were considerably
blurrier than the results of our SDA-based methods, especially for $\zeta=0.2$.

It was found that for $\dot{\epsilon}$ in the range of $\left[ 10^{-6}, 10^{-2}\right]$ worked well for PURE approximation,
which is similar to the SURE case. At high values of $\zeta > 0.1$, the approximation became highly inaccurate and  
after $\zeta > 0.2$ the model could not converge and thus training was not feasible. 
These findings about the accuracy of PURE approximation are in agreement with~\cite{le2014unbiased}.

\begin{table}[!h]
	%\vskip -0.1in
	\caption{Results of Poisson noise denoisers for MNIST (Performance in dB). Averages of 10 experiments are reported.
	}
	\label{table:MNIST_pure}
	\centering
	%	\vskip 0.15in
	\footnotesize
	{\begin{tabular}{c|cc|c}
			\toprule
			Methods 		& BM3D+VST & SDA-PURE & SDA-MSE-GT\\
			\midrule
			$\zeta = 0.01$ & \bf{30.57} & 30.55 &30.55  \\
			$\zeta = 0.1$ & 22.34 & \bf{25.80} & 25.99 \\
			$\zeta = 0.2$ & 19.56 & \bf{23.51} & 24.27 \\
			\bottomrule
	\end{tabular}}
\end{table}

\section{Discussion}
\label{discuss}

Our proposed SURE-based DNN denoisers can be useful for the
applications with large amounts of noisy images, but with few noiseless images.  %, or with expensive noiseless images. 
Note that we proposed method to train and refine DNN denoisers in unsupervised ways rather than the denoiser network itself.
Since deep learning based denoising research for novel DNN architectures is still evolving, 
incorporating our proposed SURE and SURE-FT methods into these new high performance DNN denoiser networks could possibly 
%and it may be even possible for our SURE-based training method
to achieve significantly better performances
than BM3D, or other conventional state-of-the-art denoisers. %, when it is applied to novel DNN denoisers.
Further investigation will be needed for high performance denoising networks for synthetic and real noise.

Our proposed SURE-FT method looks similar to recently proposed DIP~\cite{ulyanov2017deep} in terms of training (or refining) a DNN network for a given test image. 
While DIP has a wide variety of applications such as denoising, super resolution, inpainting, and reconstruction, our SURE-FT is restricted to Gaussian denoising.
However, as shown in this article, our proposed SURE-FT yielded significantly better performance than DIP for Gaussian noise removal.
SURE-FT can be applied to refine many existing, pretrained deep learning based denoisers, while DIP requires special network architecture such as 
U-Net~\cite{Ronneberger:2015gk} to achieve high performance.
%. When it is applied to fine-tune pretrained networks, there are practically no restrictions to the network architecture. Even pretrained networks
%containing batch normalization layers can be utilized as long as those layers are frozen during fine-tuning.
We also tried to train a network from scratch with SURE-FT (i.e. with a single noisy test image) that is similar to DIP, but
that network was not able to yield good denoised performance.
%that may require certain architectures such as 
%However, even in this case the denoising capabilites of the network may be limited since it will not 
Thus, it seems important to apply SURE-FT to pretrained networks such as CDnCNN-SURE to benefit from the information provided by large training dataset.
In other words, our SURE-FT combines both %, we 
the state-of-the-art deep learning based denoising and learning to denoise the unique structures of an input image.

In this work, Gaussian with known $\sigma$ 
and Poisson noise with known $\zeta$
was assumed in all simulations. 
However, there are several methods to estimate those parameters 
that can be used with our methods (see~\cite{Ramani:2008ij} and ~\cite{le2014unbiased}
for details).
SURE can 
incorporate a variety of noise distributions other than Gaussian noise. % with constant variance. 
%For example, SURE has been used for parameter selection of conventional
%filters for a Poisson distribution~\cite{deledalle2014stein}. 
Generalized SURE for exponential families has been proposed~\cite{Eldar:2009fga}
so that other common noise types in imaging systems can be potentially considered %used
for SURE-based methods. 
It should be noted that SURE does not require any prior knowledge on images.
Thus, %potentially %, so
it can be applied to the measurement domain for different applications,
such as medical imaging.
%. For example, 
%some measurement domain in medical imaging follows Poisson distribution (CT, PET, SPECT) and
%there are other measurement domain that contains complex Gaussian noise (MRI).
%However, especially in %image domain of 
%medical imaging, there is no known exact noise distribution in image domain.
%In order to apply our proposed methods to the image domain in medical imaging, 
Owing to noise correlation (colored noise) in the image domain 
(e.g., based on the Radon transform in the case of CT or PET), further investigations 
will be necessary to apply our proposed method directly to the image domain.
%If measurement and image domains are different ($e.g.$, Fourier, Radon transform), then
%the noise property in image domain should be carefully investigated in order to use our proposed method in image domain.
%is complex, but there have been some works
%to analyze it so that it approximately follows 
%weighted Gaussian distribution~\cite{Barrett:1994wa,Fessler:1996hu,Gravel:2004kz}.
%Since SURE expression is the sum of all unbiased risk estimator for each pixel, 
%it is trivial to apply SURE locally with spatially varying variance.
%However, 

%it will be challenging to apply SURE to image domain directly.
Note that unlike~(\ref{eq:riskdenoiser}), the existence of the minimizer for (\ref{eq:riskdenoiser_emp_batch_prop_mcsure}) should be considered with care since
it is theoretically possible that (\ref{eq:riskdenoiser_emp_batch_prop_mcsure}) becomes
negative infinity due to the divergence term in (\ref{eq:riskdenoiser_emp_batch_prop_mcsure}). 
However, in practice, this issue can be addressed by introducing a regularizer (weight decay), with
a DNN structure so that % for denoisers to have
denoisers can impose regularity conditions on the function $h$ (e.g., bounded norm of $\nabla h$),
either by choosing an adequate $\epsilon$ value,
or by using proper training data. Lastly, note that we derived 
(\ref{eq:riskdenoiser_emp_batch_prop_mcsure}), an unbiased estimator for MSE, assuming a
fixed $\boldsymbol \theta$. Thus, there is no guarantee that
the resulting estimator (denoiser) that is tuned by SURE will be unbiased~\cite{tibshirani2016excess}.

\section{Conclusion}
\label{conclude}

We proposed a SURE based training method for general deep learning denoisers in unsupervised ways.
Our proposed method with SURE trained DNN denoisers without noiseless ground truth data
so that they could yield comparable denoising performances to those elicited by
the same denoisers that were trained with noiseless ground truth data,
and outperform the conventional state-of-the-art BM3D.
Our proposed SURE-based refining method with a noisy test image further improved performance and  
outperformed conventional BM3D, deep image prior, and often the networks trained with ground truth for both grayscale and color images.
Potential extension of our SURE-based methods to Poisson noise model was also demonstrated.

\ifCLASSOPTIONcompsoc
  \section*{Acknowledgments}
  This work was supported partly by 
  Basic Science Research Program through the National Research Foundation of Korea(NRF) 
  funded by the Ministry of Education(NRF-2017R1D1A1B05035810),
  the Technology Innovation Program or Industrial Strategic Technology Development Program 
  (10077533, Development of robotic manipulation algorithm for grasping/assembling 
  with the machine learning using visual and tactile sensing information) 
  funded by the Ministry of Trade, Industry \& Energy (MOTIE, Korea), and a grant of the Korea Health Technology R\&D Project 
  through the Korea Health Industry Development Institute (KHIDI), 
  funded by the Ministry of Health \& Welfare, Republic of Korea (grant number: HI18C0316).
\else
  \section*{Acknowledgment}
\fi

\ifCLASSOPTIONcaptionsoff
  \newpage
\fi



%\bibliographystyle{IEEEtran}

\bibliographystyle{IEEEtran}
\bibliography{denoising}

\end{document}